\documentclass{article}

\PassOptionsToPackage{numbers, compress}{natbib}
\usepackage[final]{neurips_2024}

\usepackage[utf8]{inputenc} 
\usepackage[T1]{fontenc}    
\usepackage{hyperref}       
\usepackage{url}            
\usepackage{booktabs}       
\usepackage{amsfonts}       
\usepackage{nicefrac}       
\usepackage{microtype}      
\usepackage{xcolor}         
\usepackage{outlines}
\usepackage{svg}
\usepackage{mathtools}

\DeclareMathOperator*{\argminA}{arg\,min}

\newcommand{\Gauss}{\mathcal{N}}
\newcommand{\approxtext}[1]{\ensuremath{\stackrel{\text{#1}}{\approx}}}

\title{Can Bayesian Neural Networks Make Confident Predictions?}

\author{%
  Katharine Fisher \\
  Massachusetts Institute of Technology\\
  Cambridge, MA 02139 \\
  \texttt{kefisher@mit.edu} \\
   \And
   Youssef Marzouk \\
  Massachusetts Institute of Technology\\
  Cambridge, MA 02139 \\
  \texttt{ymarz@mit.edu} \\
}

\begin{document}

\maketitle

\begin{abstract}
  Bayesian inference promises a framework for principled uncertainty quantification of neural network predictions. Barriers to adoption include the difficulty of fully characterizing posterior distributions on network parameters and the interpretability of posterior predictive distributions. We demonstrate that under a discretized prior for the inner layer weights, we can exactly characterize the posterior predictive distribution as a Gaussian mixture. This setting allows us to define equivalence classes of network parameter values which produce the same likelihood (training error) and to relate the elements of these classes to the network’s scaling regime---defined via ratios of the training sample size, the size of each layer, and the number of final layer parameters. Of particular interest are distinct parameter realizations that map to low training error and yet correspond to distinct modes in the posterior predictive distribution. We identify settings that exhibit such predictive multimodality, and thus provide insight into the accuracy of unimodal posterior approximations. We also characterize the capacity of a model to ``learn from data'' by evaluating contraction of the posterior predictive in different scaling regimes.

 \end{abstract}

\section{Introduction}

Uncertainty is key to learning. Questions of how to quantify neural network prediction uncertainty are inextricable from questions of how expressive models learn to generalize \citep{Zhang2017,Zhang2021,Poggio2018}. Progress on these questions has been made through analysis of relatively simple networks, including random feature models \citep{Rahimi2007} and neural tangent kernels \citep{Jacot2018}, which demonstrate the double descent phenomenon \citep{Belkin2019,Bartlett2021,Mei2022,Clarte2023,Adlam2020TheNT}. An array of uncertainty metrics have been proposed for neural networks, as detailed by \citep{Gawlikowski2023}, but most approaches rely on heuristics which make interpretation challenging even in simple networks. 

Bayesian neural networks (BNNs) promise a principled framework for obtaining predictive distributions conditioned on training data \citep{Neal1996,Mackay1992, Arbel2023}. Realizing this promise has been complicated by the need to design appropriate prior and likelihood models and to characterize multimodal posterior distributions. Locating all modes via sampling is generally intractable, though mode connectivity and algorithm choice may aid in discovering parameter values that successfully generalize \citep{Garipov2018,pmlr-v139-izmailov21a,Rossi2024,neyshabur2015}. Many strategies for approximate inference in BNNs have also been developed. The Laplace approximation \citep{Denker1991,MacKay1992b} represents the predictive distribution with a single mode. Variational inference methods \citep{Hinton1993,Barber1998,Blei2017} are more flexible, but typically capture at most a few posterior modes. Such approaches seem to risk underestimating uncertainty, though the debate about ``cold posteriors'' has raised the possibility that narrower distributions may produce better generalization \citep{pmlr-v119-wenzel20a}.  Partially Bayesian networks \citep{Kristiadi2020, pmlr-v206-sharma23a} could offer uncertainty estimates without introducing the challenge of learning distributions over all parameters. Broadly, however, there is a need for tools that provide insight into what these approximations of the Bayesian posterior miss. 

In this work, we demonstrate that adopting a discrete prior on the inner layer weights of a BNN is a useful tool for accessing the predictive distribution without exhaustively sampling parameter space. Such priors allow us to identify cases where different posterior modes map to distinct modes in the predictive distribution. Then we can determine when predictions based on a single posterior mode will fail. To the authors' knowledge, this approach to analyzing multimodality is unique, though the different treatment of inner and final layer parameters during inference bears some resemblance to work on subnetwork inference  \citep{pmlr-v139-daxberger21a, pmlr-v206-sharma23a}, partial Bayesian networks \citep{Kristiadi2020, pmlr-v206-sharma23a}, and random feature models \citep{Rahimi2007}. Furthermore, characterizing the posterior predictive distribution allows us to identify settings where the predictive uncertainty does \textit{not} contract as the network and training set size grow proportionally. This behavior raises the question of whether overparameterized BNNs can produce ``confident predictions,'' i.e., predictions whose posterior distribution contracts around the truth as the network and data set size grow.

Section~\ref{model} outlines our model and approach to inference.  Sections~\ref{gaussian} and \ref{constructed} examine the impact of network and training set size on predictive uncertainty for a discretized Gaussian prior and then for a prior which puts mass on optimal parameter values. We put our findings in the context of related work in Section \ref{related_work}, then conclude with a discussion of the implications of multimodality for approximate inference tools and the role of Bayesian uncertainty in successful generalization.

\section{Predictive distribution of a Bayesian neural network\label{model}}

We consider an \(L\)-layer neural network in a regression setting,\footnote{Extensions to classification settings are provided in Appendix \ref{A}.}
\begin{eqnarray}
	\label{eq:y_hat}
	\hat{y}  &=&  w^\top x_L , \\
	x_\ell  &=&  \sigma( \Theta_{\ell-1}^\top x_{\ell-1}   \ - \ b_{\ell-1} ), \ \qquad \ 1<\ell\leq L, \label{eq:x} 
\end{eqnarray}
where \(x_1 \in \mathbb{R}^d\) is the network input, $\hat{y} \in \mathbb{R}$ is the output, and \(\sigma\) is a nonlinear activation function that operates component-wise. The trainable parameters include the final layer weights \(w\in\mathbb{R}^p\) and interior parameters \( \Theta \coloneqq \{ \Theta_\ell \in \mathbb{R}^{ d_\ell \times d_{\ell+1} }, \ b_\ell\in\mathbb{R}^{d_\ell} \}_{\ell=1}^{L-1}  \). Note that \(d_1=d\) and \(d_L=p\). We make the prior assumption that
\begin{eqnarray}
	w \sim \Gauss ( 0, p^{-1}\mathbf{I}_p), \qquad
    \mathbb{P}  ( \Theta = \Theta^{(j)} )  = \rho_j, \qquad \sum_{j=1}^J \rho_j \ = \ 1,
\end{eqnarray}
where each \(\Theta^{(j)}\) is a fixed realization of the interior parameters. Crucially, the discrete prior on \(\Theta\) allows us to derive an analytical representation of the Bayesian posterior predictive distribution. 

Our training set has the form $\{ ( x_1^{(i)}, y^{(i)} ) \}_{i=1}^n$ where we assume that 
\begin{eqnarray}
	y^{(i)}  =  g(x_1^{(i)}) \ + \ \varepsilon^{(i)},  \qquad \varepsilon^{(i)} \overset{\text{iid} }{\sim} & N(0, \,  \gamma^2),
\end{eqnarray}
for some (unknown) data-generating function $g: \mathbb{R}^d \to \mathbb{R}$. For convenience, we define \(X_\ell \coloneqq [x_{\ell}^{(1)},  \dots ,  x_{\ell}^{(n)} ] \in \mathbb{R}^{d_\ell \times n}\), for any $\ell \in [L]$,  and \(Y \coloneqq ( y^{(1)},  \dots ,  y^{(n)}) \in\mathbb{R}^n\).The training data can thus be written more concisely as \( (X_1, Y) \). Note that we impose no assumption on the distribution of \(X_1\). 

Let $\widetilde{x}_1 \in \mathbb{R}^d$ be an input value at which we will test our network predictions and let $\widetilde{y} \in \mathbb{R}$ denote the corresponding output. Under our model assumptions, the posterior predictive density for $\widetilde{y}$ is a $J$-component Gaussian mixture:
\begin{eqnarray}
	\label{eq:posterior_predictive}
    \pi( \widetilde{y} \, \vert \,  X_1,Y, \widetilde{x}_1  ) &=&\ \sum_{j=1}^J \mathbb{P}( \Theta^{(j)}  \, \vert \, X_1,Y )\ \pi( \widetilde{y}  \,  \vert \,  X_1, Y, \widetilde{x}_1, \Theta^{(j)} ) .
\end{eqnarray}
For each \(j\), Bayesian linear regression yields
\begin{eqnarray}
	\label{eq:y_dist}
	\pi(\widetilde{y} \, \vert \,  X_1, Y, \widetilde{x}_1, \Theta^{(j)} ) &=& 				
	\Gauss \bigl (\widetilde{y}; \,  p^{-1}\widetilde{x}_L^\top X_L ( p^{-1}X_L^\top X_L \ + \ \gamma^2 
	\mathbf{I} )^{-1} Y, \nonumber \\ 
	&&  \qquad \gamma^2   \ + \ \gamma^2 p^{-1} \widetilde{x}_L^\top  	
	( p^{-1}X_L X_L^\top \ + \ \gamma^2 \mathbf{I} )^{-1} \widetilde{x}_L   \bigr ),
\end{eqnarray}
where dependence on \(\widetilde{x}_1\) in the mean and variance terms above enters via \(\widetilde{x}_L\), as described in \eqref{eq:x}. Note that both \(X_L\) and \(\widetilde{x}_L\) depend on \(\Theta^{(j)}\). By Bayes' rule, the mixture weights are
\begin{eqnarray}
	\label{eq:weight}
    \mathbb{P}( \Theta^{(j)}  \big | X_1, Y ) \ = \  \frac{\rho_j \, \pi( Y | X_1, \Theta^{(j)} )}{\pi( Y | X_1  )}  \ = \ \left( 1 + \sum_{k\neq j} \frac{\rho_k\pi(Y | X_1, \Theta^{(k)})}{ \rho_j\pi(Y | X_1, \Theta^{(j)}) } \right)^{-1},
\end{eqnarray}
where 
\begin{eqnarray}
	\label{eq:likelihood}
	 \pi(Y| X_1, \Theta^{(j)})  = \Gauss \bigl (Y;  \mathbf{0}, \ p^{-1} X_L^T X_L + \gamma^2 \mathbf{I} \bigr )  \eqqcolon \mathcal{L}(X_L(\Theta^{(j)})) 
\end{eqnarray}
is the marginal likelihood function for  \(\Theta^{(j)}\). For simplicity, we may abbreviate
  \( \mathcal{L}(X_L(\Theta^{(j)})) \) as \( \mathcal{L}(\Theta^{(j)}) \). Assuming that \(\rho_j=1/J\) for all \(j\in[J]\), the \(j^{th}\) mode  of the posterior predictive will have a  large weight only if \( \mathcal{L}(\Theta^{(j)}) \) is large compared with the marginal likelihood of all other candidate $\Theta$ values.

\section{Multimodality under a discretized Gaussian prior\label{gaussian}}

At this stage, it is not obvious whether multimodal distributions on \((w,\Theta)\) map to multimodal distributions in the space of predictions, e.g., the distribution of $\widetilde{y}$ at at given input $\widetilde{x}_1$.\footnote{In this paper, we focus only on marginal predictive distributions. It is straightforward, however, to characterize the joint distribution of predictions at any collection of different input values.} As \eqref{eq:posterior_predictive} shows, the posterior predictive distribution is the average over predictive distributions obtained by fixing each \(\Theta^{(j)}\) and inferring \(w\). Thus, the predictive distribution can be interpreted as an average over random feature models, where the weight of the \(j^{th}\) model is determined by how compatible \(\Theta^{(j)}\) is with \(Y\) compared to each \(\Theta^{(k \neq j)}\). It is natural to ask for which regimes of \(n\), \(p\), and \(d\) it is possible to obtain multiple realizations of \(\Theta\) that each produce high marginal likelihoods \(\mathcal{L}(\Theta)\), but map to \emph{distinct} predictive modes.

In this section, we consider two-layer networks where bias parameters are set to \(0\) and the remaining components of \(\{\Theta^{(j)}\}_{j=1}^J\) are fixed by independently sampling from \(\Gauss(0, c/d )\) for some constant \(c\). We set each \(\rho_j=1/J\). Note that this choice of prior may be considered a discretization of a Gaussian prior, a common minimally informative choice for BNNs \citep{Arbel2023}. As is generally the case for Monte Carlo schemes, it is intractable to fully explore the continuous parameter space represented by a Gaussian prior, but larger \(J\) will correspond to greater coverage. For our experiments, we choose \(J=200\,000\). Appendix \ref{C} demonstrates that smaller choices of \(J\) produce similar results. The columns of \(X\) and \(\widetilde{x}_1\) are drawn from standard Gaussian distributions, and \(Y\) and \(\widetilde{y}\) are chosen to have unit variance. We consider the rectified linear unit (ReLU) activation function, and set \(c=2\pi/(\pi-1)\) so that the prior predictive distribution has unit variance.

\begin{figure*}
	\begin{tabular}{cc}
	\includegraphics[width=0.68\linewidth]{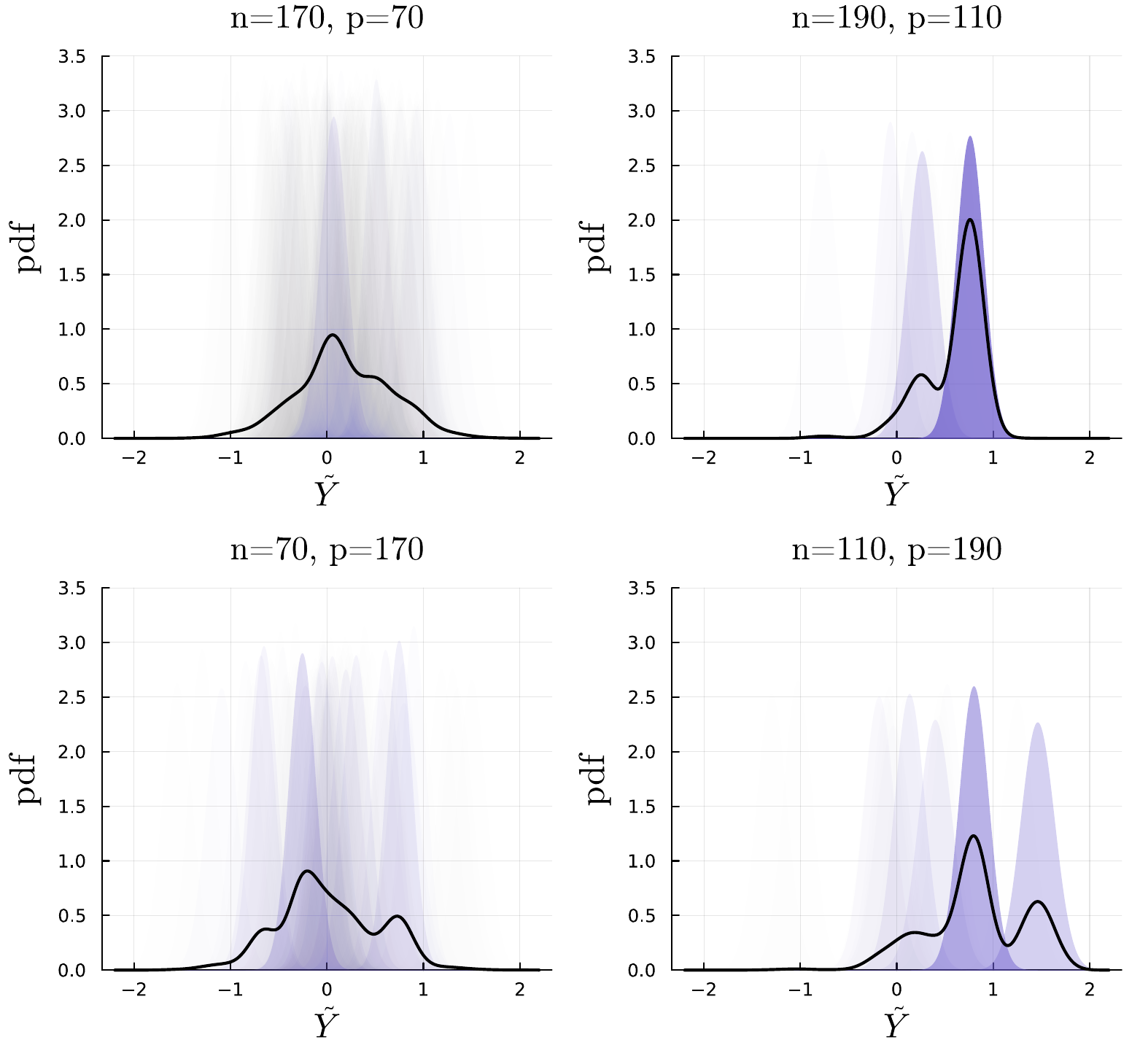} &
	\includegraphics[width=0.31\linewidth]{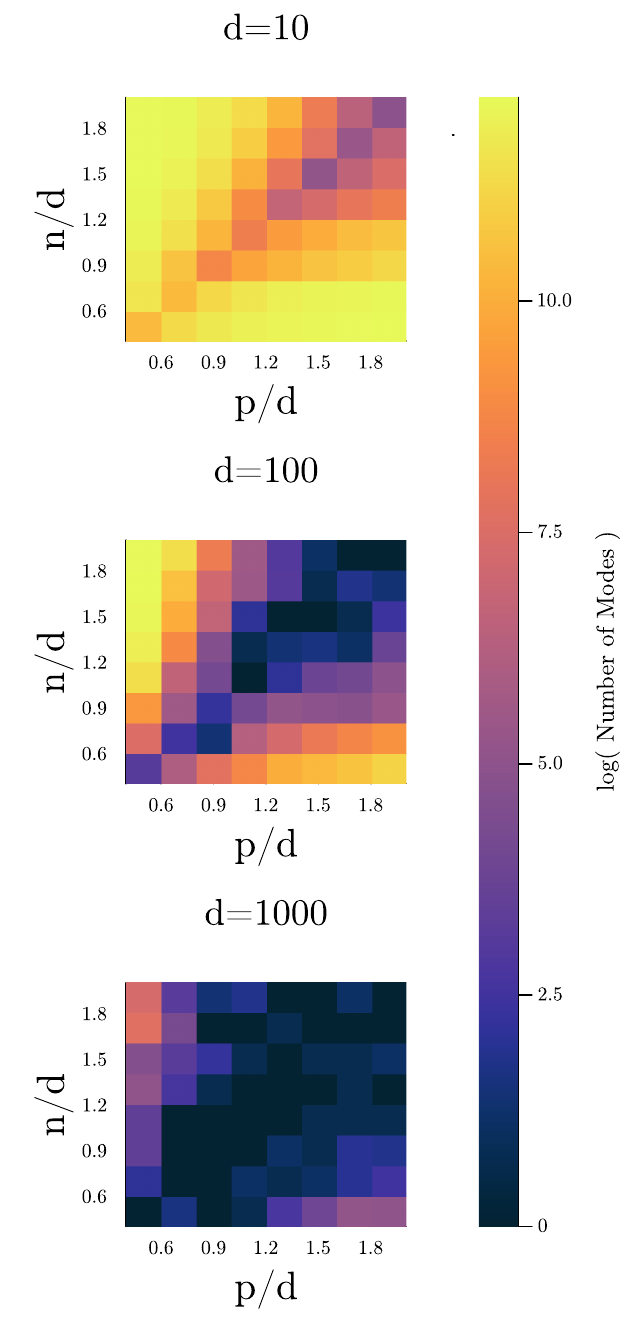}
	\end{tabular} 
    \caption{\label{fig:empirical_pdfs} Left and center: posterior predictive distributions for input dimension \(d=100\) at select training set sizes \(n\) and final layer widths \(p\), as indicated by each title. The black line shows the pdf which is a mixture of Gaussians. Each shaded distribution is a component of this mixture with transparency corresponding to its weight. Right: Heatmaps depicting the log of the number of component distributions which have weight larger than \(10^{-6}\) for specified network dimensions. Observation noise variance is set to \(\gamma^2=0.01\) for these results.}
\end{figure*}

Figure \ref{fig:empirical_pdfs} summarizes the findings of these experiments. The left and center columns show predictive distributions at a given \(\widetilde{x}_1\)
for select \( (p,n)\) pairs and \(d=100\). Each indigo region represents the predictive distribution corresponding to a candidate \(\Theta^{(j)}\); darker shades indicate larger weights as given by \eqref{eq:weight}. The black curve marks the full posterior predictive distribution. Clearly, in our setting, Bayesian inference can produce multimodal predictive distributions. Each posterior predictive distribution demonstrates smaller variance than the prior predictive distribution; thus, conditioning on training data has reduced uncertainty. Appendix \ref{B} provides examples of the posterior predictive distributions at additional test points and for larger network sizes. 

The rightmost column of Figure \ref{fig:empirical_pdfs} documents a more extensive exploration of the impact of \(n\), \(p\), and \(d\). For \(d\in\{10,100,1000\}\) and ratios \(n/d\) and \(p/d\) ranging from \(0.5\) to \(2\), we plot the log of the number of candidates \(\Theta^{(j)}\) which produce a Gaussian mixture component with weight larger than \(10^{-6}\). Note that the total number of trainable parameters in the network we consider is \(p(d+1)\), so each network considered is overparameterized. If we restrict our attention to inference in the final layer weights \(w\), however, only the entries below the right leaning diagonal of each heatmap correspond to overparameterized networks. 

Network and training set size clearly influence the number of modes that are significant in the posterior predictive distribution. When \(d=10\) we see that for several of the \(n\) and \(p\) values considered, up to 98\% of the candidate inner layer parameter values make significant contributions to the posterior predictive distribution. By contrast, for larger input dimensions, when \(n\) is close to \(p\) we often find only one significant mode, leading to a unimodal posterior predictive distribution. These findings are expected if we recall the dependence of \eqref{eq:weight} on \(\mathcal{L}(\Theta)\). If \(p\) is sufficiently large compared to \(n\), many candidates \(\Theta^{(j)}\) will produce large \(\mathcal{L}(\Theta)\) due to final layer overparameterization. Since \(X_{L-1}^\top \Theta^{(j)}\) is full rank with high probability, if \(n\) is much larger than \(p\), it becomes challenging to identify a single candidate \(\Theta^{(j)}\) which could reproduce the training data; but many of the available candidates produce similar \(\mathcal{L}(\Theta)\) and thus contribute to the posterior predictive. The line \(n=p\) represents a phase transition around which one or a few candidates are likely to outperform the others. Appendix \ref{C} provides more discussion of the impact of this transition from under- to overparameterization (in terms of final layer weights).

It is notable that the dark blue region, where few candidates produce significant modes, becomes larger as the network and training set sizes increase. Among our results, mixtures with a smaller number of component modes tend to have smaller predictive variance, as discussed in Appendix \ref{C}. This empirical observation suggests contraction of the posterior predictive as \(n\) increases. However, we also find that the range of \( \mathcal{L}(\Theta) \)
values widens with \(n\), so we can expect that as \(n\) increases, the number of candidates $J$ necessary to adequately cover parameter space will also increase.

These numerical experiments suggest that the full posterior predictive distribution often will not be well represented by an approximation that is based on a single candidate parameter value \(\Theta^{(j)}\) producing low training error, i.e., high marginal likelihood \(\mathcal{L}(\Theta^{(j)})\). Of course, our model for inference does not fully capture the predictive distribution that would be obtained with a continuous prior distribution. It is possible that if we increased \(J\) or identified candidate network parameters \(\Theta^{(j)}\) with more specific structure, we would find one dominating component of the predictive distribution, or instead see a ``filling in'' of the predictive distribution. That is, there might be components between existing components that render the continuous predictive distribution unimodal. If such a ``filling in'' occurs, however, approximations based on one particularly good candidate \(\Theta^{(j)}\) would still underestimate the true posterior uncertainty. This possibility opens questions of whether overparameterized BNNs can successfully ``forget their priors'' to learn from data, and whether a fully Bayesian model of uncertainty is suitable for producing low generalization error.
In the next section, we will contrast these initial experiments with predictive distributions found by deliberately constructing inner layer parameter candidates with greater structure.

\section{Constructing optimal parameters\label{constructed}}

\begin{figure*}
	\begin{tabular}{cc}
	\includegraphics[width=0.43\linewidth]{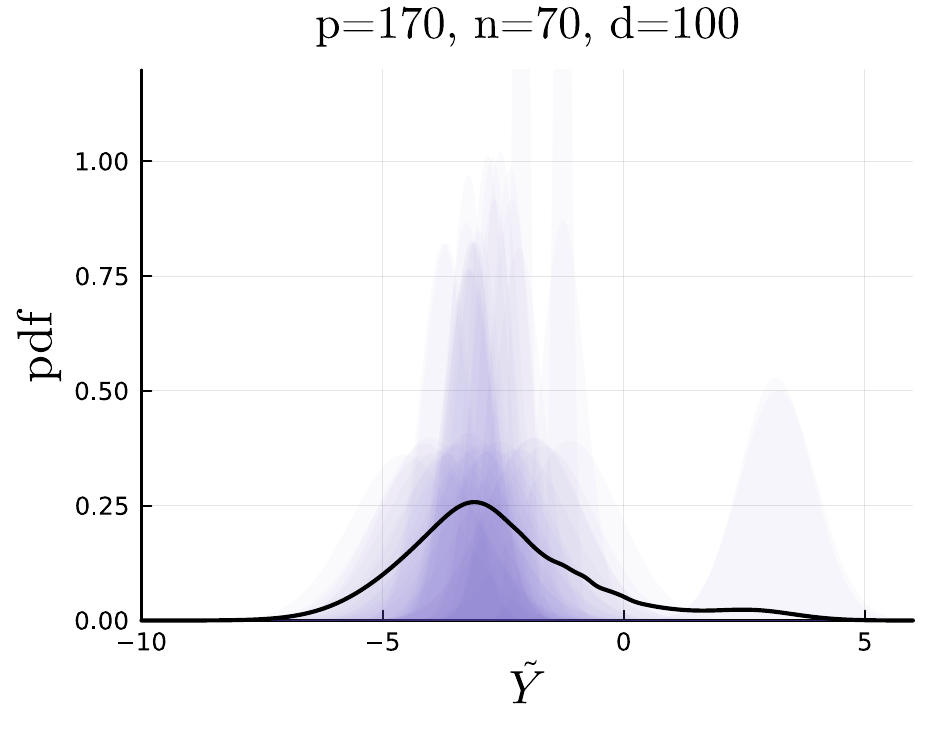} &
	\includegraphics[width=0.41\linewidth]{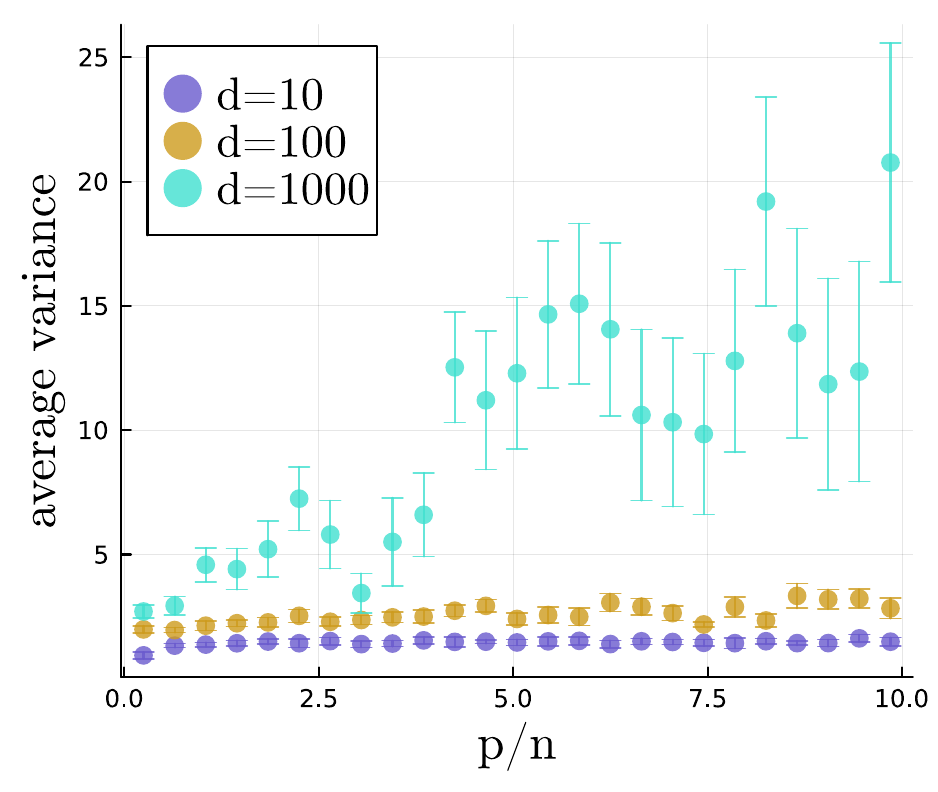}  \\
	\includegraphics[width=0.43\linewidth]{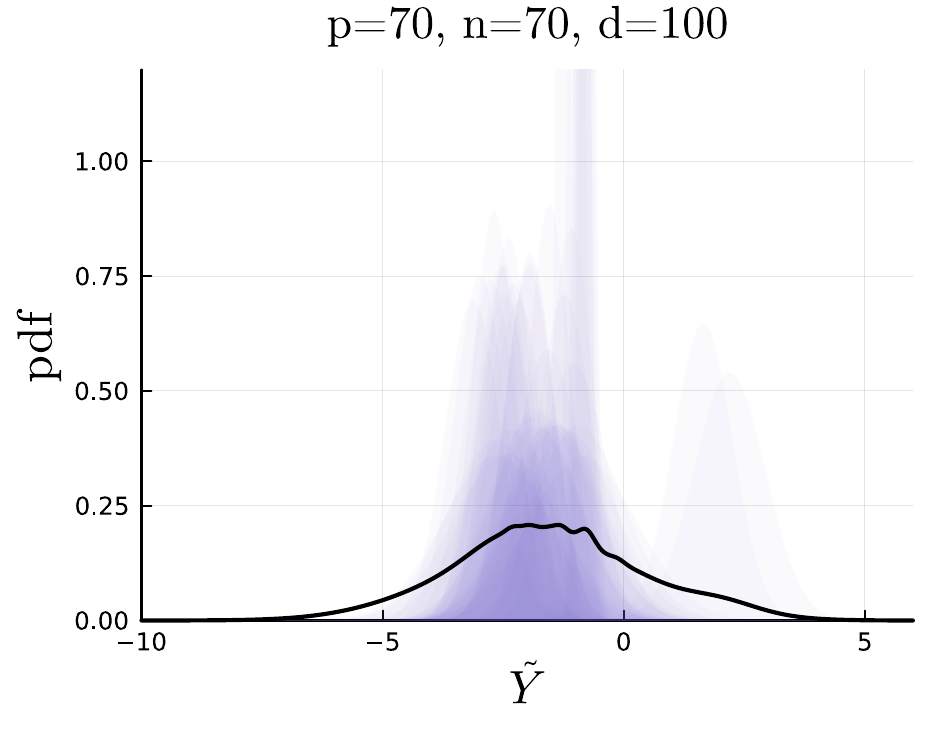} & 
	\includegraphics[width=0.43\linewidth]{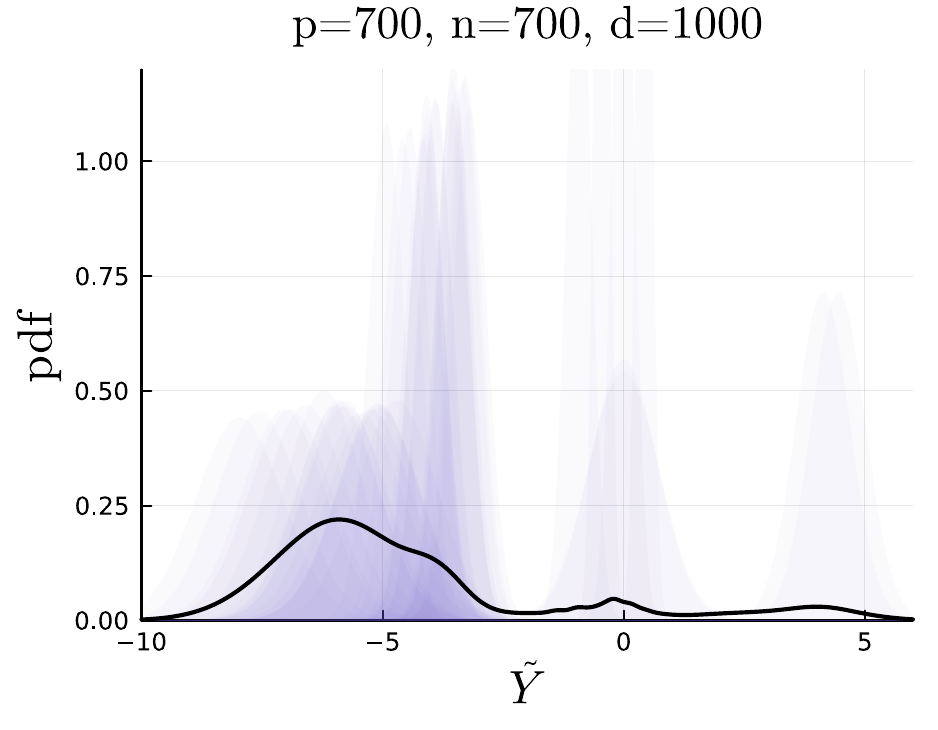} \\
	\end{tabular} 
    \caption{\label{fig:optimized_pdfs}  Top left and bottom: Predictive distributions based on candidate parameters constructed to achieve \eqref{eq:conjecture}. The full distribution is plotted in black and components are shaded according to their weight in indigo. We consider \(10\) rotations, \(10\) preimage samples, and \(10\) column space samples to construct the distribution --- a total of \(1000\) samples. Top right: The scale of predictive distributions for select \(d\) and \(p/n\) where \(n/d=0.7\). We plot the mean and standard error obtained from \(10\) realizations of \(Y\) for which we find the median predictive variance across \(10\) realizations of \(\widetilde{x}_1\).  }
\end{figure*}

As discussed above, candidate network parameters produced by drawing a finite set of samples from a Gaussian prior might omit a parameter value that would qualitatively change the behavior of the posterior predictive. To address this limitation, we can use our observation that the weight of each component of the posterior predictive distribution \eqref{eq:posterior_predictive} depends on the marginal likelihood of the corresponding candidate \eqref{eq:weight}. Now, we identify a set of candidates \(\Theta^{(j)}\) which have high marginal likelihood by construction, and show that a prior which puts mass on these candidates produces a multimodal predictive distribution. A first step toward identifying these candidates is to consider the upper bound 
\begin{eqnarray}
	\label{eq:optimization}
	\max_{\Theta} \frac{1}{n}\log\mathcal{L}(X_L(\Theta); \ X_1,Y)  \ \leq \ \max_{X_L}\frac{1}{n}\log\mathcal{L}(X_L; \ X_1,Y).
\end{eqnarray}
As detailed in Appendix \ref{D}, the matrix \(X_L^*\) solving the optimization problem on the right of \eqref{eq:optimization} satisfies 
\begin{eqnarray}
	\label{eq:optimum}
	X_L^{*\top} X_L^* \ = \ YY^\top ( 1 - \gamma^2 (Y^\top Y)^{-1} ).
\end{eqnarray}
The existence of one or more \(\Theta^*\) that map to this optimal \(X_L\) depends on the choice of activation function. If we consider ReLU activation, then all elements of \(X_L(\Theta^*)\) must be nonnegative. If the elements of \(Y\) are drawn from a centered distribution with unit variance, the probability that all elements of \eqref{eq:optimum} are nonnegative is vanishingly small. Since \(X_L^\top X_L\) is an estimate of the covariance of \(Y\), we conjecture that if we add the constraint that its elements must be nonnegative to the right hand side of \eqref{eq:optimization}, we obtain
\begin{eqnarray}
	\label{eq:conjecture}
	X_L^{*\top} X_L^* \ \approx \ \sigma(YY^\top)  ( 1 - \gamma^2 (Y^\top Y)^{-1} ) .
\end{eqnarray}
Note that we require \(n\leq d_{L-1}\) to be guaranteed a solution \(\Theta^*\) which maps to the right hand side above. We test \eqref{eq:conjecture} empirically in Appendix \ref{E}.

We may now consider an equivalence class \([\Theta]_{\mathcal{L}}\) of network parameters \(\Theta\) that map to \(X_L^{*\top} X_L^*\) as defined in \eqref{eq:conjecture}. All elements of this class will have \textit{identical} training error and marginal likelihood, \(\mathcal{L}(\Theta)\). The parameters of ReLU networks are both permutation and scale invariant \citep{rochussen2024structuredpartialstochasticitybayesian}; thus, multiple realizations of \((\Theta,w)\) map to both identical training error and identical test predictions. But it is also possible to construct \(\{\Theta^{(j)}\}_{j=1}^J\) that map to the same training error without relying on permutation and scale invariance. Specifically, we can consider unitary rotations of \(X_L\) which preserve the nonegativity of its elements, samples of the preimage of the activation function, and samples of the column space of \(X_{L-1}\). Using these constructions, we will demonstrate that not all elements of \([\Theta]_{\mathcal{L}}\) correspond to identical predictive distributions.

Figure \ref{fig:optimized_pdfs} describes posterior predictive distributions for a two-layer network with \(b_1=0\) and a prior on the inner layer parameters that puts mass on elements of \([\Theta]_{\mathcal{L}}\). For a comparison of this prior to a continuous Gaussian prior, see Appendix \ref{F}. We consider the combinatorial space of parameter candidates created from 10 rotations of \(X_2^*\), 10 samples of the preimage of ReLU, and 10 samples of the column space of \(X_1\). The top left subfigure shows the posterior predictive distribution for the same network size and \(\widetilde{x}_1\) considered in the lower left corner of Figure \ref{fig:empirical_pdfs}. As in Section \ref{gaussian}, we find examples of different candidates \(\Theta^{(j)}\) which  map to distinct predictive distributions. In contrast to Section \ref{gaussian}, we also find this behavior for networks where \(n=p\), as shown by plots on the bottom row of Figure \ref{fig:empirical_pdfs}. These results suggest that for many BNNs, the true Bayesian uncertainty of the posterior predictive distribution will be influenced by multiple modes of the posterior, moderated by the prior. 

To understand the influence of the prior, we must examine predictive variance. Large variance indicates that the training data is not sufficient to distinguish between the parameter candidates given weight by the prior. Too much prior influence may produce poor generalization. The proportional asymptotics limit---where \(n,p,d\to\infty\) while the ratios between these values remain fixed and finite---has been an important setting for examining the generalization of two-layer neural networks \citep{Mei2022,Gerace2021}. Significantly, when we consider parameter candidates from \([\Theta]_{\mathcal{L}}\), the posterior predictive distribution does \textit{not} contract as \(n\), \(d\), and \(p\) increase proportionally. This outcome is not obvious because when the network is overparameterized, increasing the raw parameter count may not increase expressivity, due to the symmetries discussed in Section \ref{symmetry}.

One example of this non-contraction is shown in the bottom row of Figure \ref{fig:optimized_pdfs}, with more examples available in Appendix \ref{F}. The top right plot in Figure \ref{fig:optimized_pdfs} summarizes the scale of the posterior predictive distributions for selected network sizes and \(n/d=0.7\). For a given ratio, \(p/n\), we see that variance increases as the network and training set sizes scale proportionally. As \(d\) gets larger, we also observe a greater rate of change in variance when \(p/n\) increases.  This behavior may occur because the number of network parameters is \(p(d+1)\), so the degree of overparameterization is increasing with \(p/n\) and \(d\). Further exploration of the impact of the prior on posterior predictive variance and generalization will be a focus of our future work. 

\section{Related work \label{related_work}}

Overparameterization creates challenges for the implementation and interpretation of Bayesian neural networks. In the classical, underparameterized setting, vague priors can be forgotten through inference, but overly restrictive (and hence misspecified) model classes can produce ``confidently wrong'' behavior \citep{Kennedy2001,Sargsyan2015,Sargsyan2018}, i.e., posterior predictives that concentrate away from the truth. We can address misspecification by considering more expressive function classes, such as neural networks, but as a consequence, multiple class members may perfectly fit the training data. In practice, many approaches to obtaining point and distribution estimates for neural networks have proven effective \citep{neyshabur2017,Foong2019InBetweenUI,Immer2021ImprovingPO,Kristiadi2020,pmlr-v139-daxberger21a}, though prediction calibration is often used to evaluate the ``effectiveness'' of uncertainty quantification. Our results suggest that the full Bayesian predictive distribution may be poorly calibrated, and further work is necessary to understand how the distributions constructed through approximate methods differ from this full distribution.

\subsection{Symmetries in parameter space\label{symmetry}}

Our work questions whether overparameterized BNNs ``forget'' their prior when network size scales with the training set size. To understand the impact of raw parameter count on the predictions of a network, it is crucial to consider the mapping from model parameters to the space of predictive functions. Several works have described symmetries in network architectures: distinct points in parameter space which map to identical predictive functions \citep{HechtNielsen1990ONTA}.  While the mechanisms which produce symmetries are fully understood for networks with smooth or saturating activation \citep{Chen1993OnTG, Fefferman1994, Krkov1994, Albertini1993}, knowledge of these mechanisms is incomplete for ReLU networks \citep{Phuong2020,rolnick2020a,grigsby2023a}, though \cite{grigsby2023a} provide a non-exhaustive summary. Further, \cite{rolnick2020a} demonstrate that it is possible to reverse engineer a ReLU network up to symmetries by querying its outputs. Network symmetries motivate the definition of \emph{functional dimension}, the degrees of freedom induced in function space by small perturbations at some location in parameter space \citep{grigsby2022,grigsby2023a}. This line of work suggests that function space may be easier to explore than parameter space \citep{pmlr-v139-izmailov21a} or that parameter space may become easier to explore when symmetries are removed \citep{ainsworth2023,rochussen2024structuredpartialstochasticitybayesian,gelberg2024variational}.

Symmetries have also driven the \emph{linear mode connectivity} hypothesis, which asserts that when symmetries are taken into account, (nearly) all point estimates obtained by stochastic gradient descent (SGD) lie in the same loss basin \citep{Garipov2018,Entezari2022,ainsworth2023,Rossi2024}. The evidence towards this hypothesis has implications for the behavior of trained networks: flat basins in loss space are associated with simple solutions which satisfy Occam's razor \citep{Hochreiter94}. The work we present in this paper targets functions which are identical at training data points, but may be distinct elsewhere. In our setting, we find multiple distinct predictive modes, indicating that parameter symmetries are not the sole source of parameter space multimodality. It remains to determine whether the modes we find lie in the same loss basin once symmetries are considered, but in that case, our results would suggest that this basin contains non-trivial functional diversity.

\subsection{Modeling generalization and predictive distributions}

Techniques developed within statistical physics---particularly the replica method and Gaussian equivalence---offer models of neural network generalization error which sidestep the need for posterior sampling \citep{Mei2022,Gerace2021,Hu2020,Goldt2020}. For instance, \citep{Li2020} marginalize over parameters in a deep linear network, working from the outer layer inward to compute the Bayesian model evidence and to characterize the bias and variance of the trained network under new data. Their separate treatment of different parameter layers bears some similarity to our approach, though we consider nonlinear networks and integrate with respect to the posterior distribution rather than the prior distribution. Further, our work investigates the Bayesian predictive distribution. \citep{Hanin2023F} use Meijer G-functions to represent the characteristic function of the predictive distribution of deep linear models and demonstrate that this distribution is Gaussian in certain asymptotic regimes. While deep linear models provide useful insight into extrapolation and feature learning \citep{Li2020,ZavatoneVeth2022ContrastingRA}, these networks will be misspecified for many data-generating processes. Thus, we can expect that overparameterization has a different impact on learning with these networks compared with nonlinear networks, where only a single hidden layer is necessary to obtain a universal approximator for continuous functions \citep{Cybenko1989ApproximationBS,Funahashi1989OnTA,Hornik1989MultilayerFN}. 

An extension of Gaussian equivalence to deep nonlinear networks has recently been conjectured in \citep{Cui2023}, but that work focuses on generalization error in an idealized setting where Bayesian inference is guaranteed to obtain optimal error. Specifically, the error is averaged across data-generating networks with parameters drawn from a Gaussian, the same distribution family as the prior. Input covariates are also assumed to be standard Gaussian. Our approach can be used to study the posterior predictive obtained from arbitrary covariate distributions and data-generating models. Investigations of these cases will provide insight into whether approximations of the full Bayesian posterior introduce a trade-off between accuracy and calibration, of the kind that \citep{Clarte2023} observe for approximate methods applied to random feature models. 

\subsection{Designing predictive distributions}

In practical settings, the true predictive distribution is intractable, but successful substitutes can be constructed via approximate or heuristic methods. The definition of ``success'' may hinge on posterior consistency in the large data limit 	\citep{bhattacharya2020,wang05a,cheriefabdellatif2019}, which leaves open many questions about the behavior of the true and approximate posteriors in the overparameterized regimes. Success could also mean a balance between good accuracy and intuitive uncertainty \citep{Foong2019InBetweenUI,Kristiadi2020,pmlr-v161-kristiadi21a,Immer2021ImprovingPO}: predictive distributions which concentrate on Occam's razor solutions in the interpolatory regime and remain diffuse in the extrapolatory regime. Based on the latter notion of success, many approximate methods outperform more faithful representations of the Bayesian predictive distribution. For instance, use of the Laplace approximation and a linearized predictive model fixes the overly diffuse predictive distributions obtained by consistent Monte Carlo approximation of the posterior predictive and other implementations of the Laplace approximation \citep{Immer2021ImprovingPO}. Similarly, Laplace approximations based on hidden uncertainty units \citep{pmlr-v161-kristiadi21a} and inference in
partially stochastic networks \citep{pmlr-v139-daxberger21a,pmlr-v206-sharma23a} offer tractable methods with high accuracy and low calibration error.

These types of approximations to the posterior use only a subset of the parameters of the neural network to construct a predictive distribution. Then, it is not surprising if these aproximate distributions do not suffer from the diffuse uncertainty produced when Bayesian treatment is applied to all parameters of an overparameterized model. \citep{rochussen2024structuredpartialstochasticitybayesian} argue that partial stochasticity may be successful because it implicitly removes parameter symmetries, though we question whether the approach may also remove non-symmetric solutions with identical performance on the training data, of the kind we identify in Section~\ref{constructed}. \citep{pmlr-v206-sharma23a} establish that certain partially stochastic architectures are universal conditional distribution approximators, so limiting the parameters used to construct the predictive distribution does not necessarily limit expressivity.  A future application of our use of discrete priors to access predictive distributions may be to determine whether the distributions produced by partially stochastic networks diverge significantly from 
the full Bayesian predictive distribution. An additional piece of this puzzle is the success of cold posteriors. \citep{pmlr-v119-wenzel20a} find that sharpened posteriors perform well, while full Bayesian posteriors yield systematically worse predictions than SGD point estimates. The authors suggest that Gaussian priors and network capacity may be responsible for the underwhelming performance of the Bayesian approach. These results---alongside our own---suggest that while overparameterization is key to the successful generalization of neural networks, it is a hindrance to the construction of intuitive representations of uncertainty.

\subsection{How do we make confident predictions with BNNs?}

Bayesian inference promises principled uncertainty quantification, but we have shown that predictive distributions do not contract in overparameterized settings. Many approximate methods construct predictive distributions which perform better than fully Bayesian predictive distributions \citep{Immer2021ImprovingPO,pmlr-v119-wenzel20a}. These predictive distributions might use more information than the typical Bayesian neural network information. This information may be precisely the structural qualities which make neural networks trained with SGD successful. Implicit regularization \citep{neyshabur2017} and self-induced regularization \citep{Bartlett2021} are two examples from our evolving understanding of generalization. Future work on Bayesian inference in overparameterized models might investigate how to formally capture such structural information. 

\section{Key implications}

We have provided insight into the predictive uncertainty of Bayesian neural networks by choosing a continuous Gaussian prior for the final layer weights and a discrete prior for the interior parameters. The key implications are:
\begin{itemize}
	\item \textbf{Much of the mass of the posterior predictive distribution can be captured without sampling the entire parameter space.} For a given prior, we can construct parameter candidates with high marginal likelihood and prior weight. 
	\item \textbf{Unimodal posterior approximations are overconfident.} Multiple posterior modes contribute to the posterior predictive uncertainty of Bayesian neural networks.
	\item \textbf{The posterior predictive distribution does not contract as $n$, $p$, and $d$ increase proportionally.} Thus, in overparameterized networks, predictive uncertainty likely reflects an inability to completely forget the prior given the training data---that is, an inability to make confident predictions. 
\end{itemize}
Future work will target each of these implications. More extensive numerical experiments alongside theoretical results will consider different prior assumptions and establish minimum rates at which network size must grow with respect to training set size such that the predictive distribution does not contract. Further, we will characterize equivalence classes of parameters which map to large marginal likelihood for more network and training set sizes. Finally, we will quantify the impact of predictive uncertainty on generalization error. 

\section*{Acknowledgements}

This material is based upon work supported by the Department of Energy (DOE), National Nuclear Security Administration, PSAAP-III program, under award number DE-NA0003965. KEF also acknowledges support from the NSF Graduate Research Fellowship program, grant number 1745302. YMM also acknowledges support from the DOE Office of Science, Office of Advanced Scientific Computing Research (ASCR), under award number DE-SC0023187.

\bibliographystyle{plainnat}
\bibliography{references}

\begin{thebibliography}{64}
\providecommand{\natexlab}[1]{#1}
\providecommand{\url}[1]{\texttt{#1}}
\expandafter\ifx\csname urlstyle\endcsname\relax
  \providecommand{\doi}[1]{doi: #1}\else
  \providecommand{\doi}{doi: \begingroup \urlstyle{rm}\Url}\fi

\bibitem[Adlam and Pennington(2020)]{Adlam2020TheNT}
Ben Adlam and Jeffrey Pennington.
\newblock The neural tangent kernel in high dimensions: Triple descent and a
  multi-scale theory of generalization.
\newblock \emph{ArXiv}, abs/2008.06786, 2020.
\newblock URL \url{https://api.semanticscholar.org/CorpusID:221082525}.

\bibitem[Ainsworth et~al.(2023)Ainsworth, Hayase, and Srinivasa]{ainsworth2023}
Samuel~K. Ainsworth, Jonathan Hayase, and Siddhartha Srinivasa.
\newblock Git re-basin: Merging models modulo permutation symmetries, 2023.
\newblock URL \url{https://arxiv.org/abs/2209.04836}.

\bibitem[Albertini and Sontag(1993)]{Albertini1993}
Francesca Albertini and Eduardo~D. Sontag.
\newblock For neural networks, function determines form.
\newblock \emph{Neural Networks}, 6\penalty0 (7):\penalty0 975--990, 1993.
\newblock ISSN 0893-6080.
\newblock \doi{https://doi.org/10.1016/S0893-6080(09)80007-5}.
\newblock URL
  \url{https://www.sciencedirect.com/science/article/pii/S0893608009800075}.

\bibitem[Arbel et~al.(2023)Arbel, Pitas, Vladimirova, and Fortuin]{Arbel2023}
Julyan Arbel, Konstantinos Pitas, Mariia Vladimirova, and Vincent Fortuin.
\newblock A primer on bayesian neural networks: Review and debates.
\newblock \emph{ArXiv}, abs/2309.16314, 2023.
\newblock URL \url{https://api.semanticscholar.org/CorpusID:263134168}.

\bibitem[Barber and Bishop(1998)]{Barber1998}
David Barber and Charles~M. Bishop.
\newblock Ensemble learning in bayesian neural networks.
\newblock 1998.
\newblock URL \url{https://api.semanticscholar.org/CorpusID:14932413}.

\bibitem[Bartlett et~al.(2021)Bartlett, Montanari, and Rakhlin]{Bartlett2021}
Peter~L. Bartlett, Andrea Montanari, and Alexander Rakhlin.
\newblock Deep learning: a statistical viewpoint.
\newblock \emph{Acta Numerica}, 30:\penalty0 87–201, 2021.
\newblock \doi{10.1017/S0962492921000027}.

\bibitem[Belkin et~al.(2019)Belkin, Hsu, Ma, and Mandal]{Belkin2019}
Mikhail Belkin, Daniel Hsu, Siyuan Ma, and Soumik Mandal.
\newblock Reconciling modern machine-learning practice and the classical
  bias–variance trade-off.
\newblock \emph{Proceedings of the National Academy of Sciences}, 116\penalty0
  (32):\penalty0 15849--15854, 2019.
\newblock \doi{10.1073/pnas.1903070116}.
\newblock URL \url{https://www.pnas.org/doi/abs/10.1073/pnas.1903070116}.

\bibitem[Bhattacharya et~al.(2020)Bhattacharya, Liu, and
  Maiti]{bhattacharya2020}
Shrijita Bhattacharya, Zihuan Liu, and Tapabrata Maiti.
\newblock Variational bayes neural network: Posterior consistency,
  classification accuracy and computational challenges, 2020.
\newblock URL \url{https://arxiv.org/abs/2011.09592}.

\bibitem[Chen et~al.(1993)Chen, minn Lu, and Hecht-Nielsen]{Chen1993OnTG}
An~Mei Chen, Haw minn Lu, and Robert Hecht-Nielsen.
\newblock On the geometry of feedforward neural network error surfaces.
\newblock \emph{Neural Computation}, 5:\penalty0 910--927, 1993.
\newblock URL \url{https://api.semanticscholar.org/CorpusID:44856417}.

\bibitem[Chérief-Abdellatif(2019)]{cheriefabdellatif2019}
Badr-Eddine Chérief-Abdellatif.
\newblock Convergence rates of variational inference in sparse deep learning,
  2019.
\newblock URL \url{https://arxiv.org/abs/1908.04847}.

\bibitem[Clarte et~al.(2023)Clarte, Loureiro, Krzakala, and
  Zdeborová]{Clarte2023}
Lucas~Andry Clarte, Bruno Loureiro, Florent Krzakala, and Lenka Zdeborová.
\newblock On double-descent in uncertainty quantification in overparametrized
  models.
\newblock volume 206, page 7089–7125. PMLR Proceedings of Machine Learning
  Research, 2023.
\newblock URL \url{https://infoscience.epfl.ch/handle/20.500.14299/197303}.

\bibitem[Cui et~al.(2023)Cui, Krzakala, and Zdeborov]{Cui2023}
Hugo Cui, Florent Krzakala, and Lenka Zdeborov.
\newblock Bayes-optimal learning of deep random networks of extensive-width.
\newblock In \emph{Proceedings of the 40th International Conference on Machine
  Learning}, ICML'23. JMLR.org, 2023.

\bibitem[Cybenko(1989)]{Cybenko1989ApproximationBS}
George~V. Cybenko.
\newblock Approximation by superpositions of a sigmoidal function.
\newblock \emph{Mathematics of Control, Signals and Systems}, 2:\penalty0
  303--314, 1989.
\newblock URL \url{https://api.semanticscholar.org/CorpusID:3958369}.

\bibitem[David M.~Blei and McAuliffe(2017)]{Blei2017}
Alp~Kucukelbir David M.~Blei and Jon~D. McAuliffe.
\newblock Variational inference: A review for statisticians.
\newblock \emph{Journal of the American Statistical Association}, 112\penalty0
  (518):\penalty0 859--877, 2017.
\newblock \doi{10.1080/01621459.2017.1285773}.
\newblock URL \url{https://doi.org/10.1080/01621459.2017.1285773}.

\bibitem[Daxberger et~al.(2021)Daxberger, Nalisnick, Allingham, Antoran, and
  Hernandez-Lobato]{pmlr-v139-daxberger21a}
Erik Daxberger, Eric Nalisnick, James~U Allingham, Javier Antoran, and
  Jose~Miguel Hernandez-Lobato.
\newblock Bayesian deep learning via subnetwork inference.
\newblock In Marina Meila and Tong Zhang, editors, \emph{Proceedings of the
  38th International Conference on Machine Learning}, volume 139 of
  \emph{Proceedings of Machine Learning Research}, pages 2510--2521. PMLR,
  18--24 Jul 2021.
\newblock URL \url{https://proceedings.mlr.press/v139/daxberger21a.html}.

\bibitem[Denker and LeCun(1990)]{Denker1991}
John~S. Denker and Yann LeCun.
\newblock Transforming neural-net output levels to probability distributions.
\newblock In \emph{Proceedings of the 3rd International Conference on Neural
  Information Processing Systems}, NIPS'90, page 853–859, San Francisco, CA,
  USA, 1990. Morgan Kaufmann Publishers Inc.
\newblock ISBN 1558601848.

\bibitem[Entezari et~al.(2022)Entezari, Sedghi, Saukh, and
  Neyshabur]{Entezari2022}
Rahim Entezari, Hanie Sedghi, Olga Saukh, and Behnam Neyshabur.
\newblock The role of permutation invariance in linear mode connectivity of
  neural networks, 2022.
\newblock URL \url{https://arxiv.org/abs/2110.06296}.

\bibitem[Fefferman(1994)]{Fefferman1994}
Charles Fefferman.
\newblock Reconstructing a neural net from its output.
\newblock \emph{Revista Matematica Iberoamericana}, 10:\penalty0 507--555,
  1994.
\newblock URL \url{https://api.semanticscholar.org/CorpusID:121350232}.

\bibitem[Foong et~al.(2019)Foong, Li, Hern{\'a}ndez-Lobato, and
  Turner]{Foong2019InBetweenUI}
Andrew Y.~K. Foong, Yingzhen Li, Jos{\'e}~Miguel Hern{\'a}ndez-Lobato, and
  Richard~E. Turner.
\newblock 'in-between' uncertainty in bayesian neural networks.
\newblock \emph{ArXiv}, abs/1906.11537, 2019.
\newblock URL \url{https://api.semanticscholar.org/CorpusID:195700200}.

\bibitem[Garipov et~al.(2018)Garipov, Izmailov, Podoprikhin, Vetrov, and
  Wilson]{Garipov2018}
Timur Garipov, Pavel Izmailov, Dmitrii Podoprikhin, Dmitry Vetrov, and
  Andrew~Gordon Wilson.
\newblock Loss surfaces, mode connectivity, and fast ensembling of dnns.
\newblock In \emph{Proceedings of the 32nd International Conference on Neural
  Information Processing Systems}, NIPS'18, page 8803–8812, Red Hook, NY,
  USA, 2018. Curran Associates Inc.

\bibitem[Gawlikowski et~al.(2023)Gawlikowski, Tassi, Ali, Lee, Humt, Feng,
  Kruspe, Triebel, Jung, Roscher, Shahzad, Yang, Bamler, and
  Zhu]{Gawlikowski2023}
Jakob Gawlikowski, Cedrique Rovile~Njieutcheu Tassi, Mohsin Ali, Jongseok Lee,
  Matthias Humt, Jianxiang Feng, Anna Kruspe, Rudolph Triebel, Peter Jung,
  Ribana Roscher, Muhammad Shahzad, Wen Yang, Richard Bamler, and Xiao~Xiang
  Zhu.
\newblock A survey of uncertainty in deep neural networks.
\newblock \emph{Artif. Intell. Rev.}, 56\penalty0 (Suppl 1):\penalty0
  1513–1589, jul 2023.
\newblock ISSN 0269-2821.
\newblock \doi{10.1007/s10462-023-10562-9}.
\newblock URL \url{https://doi.org/10.1007/s10462-023-10562-9}.

\bibitem[Gelberg et~al.(2024)Gelberg, van~der Ouderaa, van~der Wilk, and
  Gal]{gelberg2024variational}
Yoav Gelberg, Tycho F.~A. van~der Ouderaa, Mark van~der Wilk, and Yarin Gal.
\newblock Variational inference failures under model symmetries: Permutation
  invariant posteriors for bayesian neural networks.
\newblock In \emph{ICML 2024 Workshop on Geometry-grounded Representation
  Learning and Generative Modeling}, 2024.
\newblock URL \url{https://openreview.net/forum?id=VCVnhR4x4v}.

\bibitem[Gerace et~al.(2021)Gerace, Loureiro, Krzakala, Mézard, and
  Zdeborová]{Gerace2021}
Federica Gerace, Bruno Loureiro, Florent Krzakala, Marc Mézard, and Lenka
  Zdeborová.
\newblock Generalisation error in learning with random features and the hidden
  manifold model*.
\newblock \emph{Journal of Statistical Mechanics: Theory and Experiment},
  2021\penalty0 (12):\penalty0 124013, dec 2021.
\newblock \doi{10.1088/1742-5468/ac3ae6}.
\newblock URL \url{https://dx.doi.org/10.1088/1742-5468/ac3ae6}.

\bibitem[Goldt et~al.(2020)Goldt, Loureiro, Reeves, Krzakala, M'ezard, and
  Zdeborov'a]{Goldt2020}
Sebastian Goldt, Bruno Loureiro, Galen Reeves, Florent Krzakala, Marc M'ezard,
  and Lenka Zdeborov'a.
\newblock The gaussian equivalence of generative models for learning with
  shallow neural networks.
\newblock In \emph{Mathematical and Scientific Machine Learning}, 2020.
\newblock URL \url{https://api.semanticscholar.org/CorpusID:235165686}.

\bibitem[Grigsby et~al.(2023)Grigsby, Lindsey, and Rolnick]{grigsby2023a}
Elisenda Grigsby, Kathryn Lindsey, and David Rolnick.
\newblock Hidden symmetries of {R}e{LU} networks.
\newblock In Andreas Krause, Emma Brunskill, Kyunghyun Cho, Barbara Engelhardt,
  Sivan Sabato, and Jonathan Scarlett, editors, \emph{Proceedings of the 40th
  International Conference on Machine Learning}, volume 202 of
  \emph{Proceedings of Machine Learning Research}, pages 11734--11760. PMLR,
  23--29 Jul 2023.
\newblock URL \url{https://proceedings.mlr.press/v202/grigsby23a.html}.

\bibitem[Grigsby et~al.(2022)Grigsby, Lindsey, Meyerhoff, and Wu]{grigsby2022}
J.~Elisenda Grigsby, Kathryn Lindsey, Robert Meyerhoff, and Chenxi Wu.
\newblock Functional dimension of feedforward relu neural networks, 2022.
\newblock URL \url{https://arxiv.org/abs/2209.04036}.

\bibitem[Hanin and Zlokapa(2023)]{Hanin2023F}
Boris Hanin and Alexander Zlokapa.
\newblock Bayesian interpolation with deep linear networks.
\newblock \emph{Proceedings of the National Academy of Sciences}, 120\penalty0
  (23):\penalty0 e2301345120, 2023.
\newblock \doi{10.1073/pnas.2301345120}.
\newblock URL \url{https://www.pnas.org/doi/abs/10.1073/pnas.2301345120}.

\bibitem[Hecht-Nielsen(1990)]{HechtNielsen1990ONTA}
Robert Hecht-Nielsen.
\newblock On the algebraic structure of feedforward network weight spaces.
\newblock 1990.
\newblock URL \url{https://api.semanticscholar.org/CorpusID:115619723}.

\bibitem[Hinton and van Camp(1993)]{Hinton1993}
Geoffrey~E. Hinton and Drew van Camp.
\newblock Keeping the neural networks simple by minimizing the description
  length of the weights.
\newblock In \emph{Proceedings of the Sixth Annual Conference on Computational
  Learning Theory}, COLT '93, page 5–13, New York, NY, USA, 1993. Association
  for Computing Machinery.
\newblock ISBN 0897916115.
\newblock \doi{10.1145/168304.168306}.
\newblock URL \url{https://doi.org/10.1145/168304.168306}.

\bibitem[Hochreiter and Schmidhuber(1994)]{Hochreiter94}
Sepp Hochreiter and J\"{u}rgen Schmidhuber.
\newblock Simplifying neural nets by discovering flat minima.
\newblock In G.~Tesauro, D.~Touretzky, and T.~Leen, editors, \emph{Advances in
  Neural Information Processing Systems}, volume~7. MIT Press, 1994.
\newblock URL
  \url{https://proceedings.neurips.cc/paper_files/paper/1994/file/01882513d5fa7c329e940dda99b12147-Paper.pdf}.

\bibitem[Hornik et~al.(1989)Hornik, Stinchcombe, and
  White]{Hornik1989MultilayerFN}
Kurt Hornik, Maxwell~B. Stinchcombe, and Halbert~L. White.
\newblock Multilayer feedforward networks are universal approximators.
\newblock \emph{Neural Networks}, 2:\penalty0 359--366, 1989.
\newblock URL \url{https://api.semanticscholar.org/CorpusID:2757547}.

\bibitem[Hu and Lu(2020)]{Hu2020}
Hong Hu and Yue~M. Lu.
\newblock Universality laws for high-dimensional learning with random features.
\newblock \emph{IEEE Transactions on Information Theory}, 69:\penalty0
  1932--1964, 2020.
\newblock URL \url{https://api.semanticscholar.org/CorpusID:221738950}.

\bibitem[ichi Funahashi(1989)]{Funahashi1989OnTA}
Ken ichi Funahashi.
\newblock On the approximate realization of continuous mappings by neural
  networks.
\newblock \emph{Neural Networks}, 2:\penalty0 183--192, 1989.
\newblock URL \url{https://api.semanticscholar.org/CorpusID:10203109}.

\bibitem[Immer et~al.(2021)Immer, Korzepa, and Bauer]{Immer2021ImprovingPO}
Alexander Immer, Maciej~Jan Korzepa, and M.~Bauer.
\newblock Improving predictions of bayesian neural nets via local
  linearization.
\newblock In \emph{International Conference on Artificial Intelligence and
  Statistics}, 2021.
\newblock URL \url{https://api.semanticscholar.org/CorpusID:221172984}.

\bibitem[Izmailov et~al.(2021)Izmailov, Vikram, Hoffman, and
  Wilson]{pmlr-v139-izmailov21a}
Pavel Izmailov, Sharad Vikram, Matthew~D Hoffman, and Andrew Gordon~Gordon
  Wilson.
\newblock What are bayesian neural network posteriors really like?
\newblock In Marina Meila and Tong Zhang, editors, \emph{Proceedings of the
  38th International Conference on Machine Learning}, volume 139 of
  \emph{Proceedings of Machine Learning Research}, pages 4629--4640. PMLR,
  18--24 Jul 2021.

\bibitem[Jacot et~al.(2018)Jacot, Gabriel, and Hongler]{Jacot2018}
Arthur Jacot, Franck Gabriel, and Cl\'{e}ment Hongler.
\newblock Neural tangent kernel: convergence and generalization in neural
  networks.
\newblock In \emph{Proceedings of the 32nd International Conference on Neural
  Information Processing Systems}, NIPS'18, page 8580–8589, Red Hook, NY,
  USA, 2018. Curran Associates Inc.

\bibitem[Kennedy and O’Hagan(2001)]{Kennedy2001}
Marc~C. Kennedy and Anthony O’Hagan.
\newblock Bayesian calibration of computer models.
\newblock \emph{Journal of the Royal Statistical Society: Series B (Statistical
  Methodology)}, 63, 2001.
\newblock URL \url{https://api.semanticscholar.org/CorpusID:119562136}.

\bibitem[Kristiadi et~al.(2020)Kristiadi, Hein, and Hennig]{Kristiadi2020}
Agustinus Kristiadi, Matthias Hein, and Philipp Hennig.
\newblock Being bayesian, even just a bit, fixes overconfidence in relu
  networks.
\newblock In \emph{Proceedings of the 37th International Conference on Machine
  Learning}, ICML'20. JMLR.org, 2020.

\bibitem[Kristiadi et~al.(2021)Kristiadi, Hein, and
  Hennig]{pmlr-v161-kristiadi21a}
Agustinus Kristiadi, Matthias Hein, and Philipp Hennig.
\newblock Learnable uncertainty under laplace approximations.
\newblock In Cassio de~Campos and Marloes~H. Maathuis, editors,
  \emph{Proceedings of the Thirty-Seventh Conference on Uncertainty in
  Artificial Intelligence}, volume 161 of \emph{Proceedings of Machine Learning
  Research}, pages 344--353. PMLR, 27--30 Jul 2021.
\newblock URL \url{https://proceedings.mlr.press/v161/kristiadi21a.html}.

\bibitem[Kůrkov{\'a} and Kainen(1994)]{Krkov1994}
Věra Kůrkov{\'a} and Paul~C. Kainen.
\newblock Functionally equivalent feedforward neural networks.
\newblock \emph{Neural Computation}, 6:\penalty0 543--558, 1994.
\newblock URL \url{https://api.semanticscholar.org/CorpusID:31012377}.

\bibitem[Li and Sompolinsky(2020)]{Li2020}
Qianyi Li and Haim Sompolinsky.
\newblock Statistical mechanics of deep linear neural networks: The
  back-propagating renormalization group.
\newblock \emph{ArXiv}, abs/2012.04030, 2020.
\newblock URL \url{https://api.semanticscholar.org/CorpusID:227745095}.

\bibitem[Lu et~al.(2021)Lu, Ie, and Sha]{lu2021}
Zhiyun Lu, Eugene Ie, and Fei Sha.
\newblock Mean-field approximation to gaussian-softmax integral with
  application to uncertainty estimation, 2021.
\newblock URL \url{https://arxiv.org/abs/2006.07584}.

\bibitem[MacKay(1992{\natexlab{a}})]{MacKay1992a}
David J.~C. MacKay.
\newblock The evidence framework applied to classification networks.
\newblock \emph{Neural Computation}, 4\penalty0 (5):\penalty0 720--736,
  1992{\natexlab{a}}.
\newblock \doi{10.1162/neco.1992.4.5.720}.

\bibitem[MacKay(1992{\natexlab{b}})]{MacKay1992b}
David J.~C. MacKay.
\newblock A practical bayesian framework for backpropagation networks.
\newblock \emph{Neural Computation}, 4\penalty0 (3):\penalty0 448--472,
  1992{\natexlab{b}}.
\newblock \doi{10.1162/neco.1992.4.3.448}.

\bibitem[MacKay(1992{\natexlab{c}})]{Mackay1992}
David J.~C. MacKay.
\newblock {Bayesian Interpolation}.
\newblock \emph{Neural Computation}, 4\penalty0 (3):\penalty0 415--447, 05
  1992{\natexlab{c}}.
\newblock ISSN 0899-7667.
\newblock \doi{10.1162/neco.1992.4.3.415}.
\newblock URL \url{https://doi.org/10.1162/neco.1992.4.3.415}.

\bibitem[Mei and Montanari(2022)]{Mei2022}
Song Mei and Andrea Montanari.
\newblock The generalization error of random features regression: Precise
  asymptotics and the double descent curve.
\newblock \emph{Communications on Pure and Applied Mathematics}, 75\penalty0
  (4):\penalty0 667--766, 2022.
\newblock \doi{https://doi.org/10.1002/cpa.22008}.
\newblock URL \url{https://onlinelibrary.wiley.com/doi/abs/10.1002/cpa.22008}.

\bibitem[Neal(1996)]{Neal1996}
Radford~M. Neal.
\newblock \emph{Bayesian Learning for Neural Networks}.
\newblock Springer, New York, 1996.
\newblock \doi{https://doi.org/10.1007/978-1-4612-0745-0}.

\bibitem[Neyshabur(2017)]{neyshabur2017}
Behnam Neyshabur.
\newblock Implicit regularization in deep learning, 2017.
\newblock URL \url{https://arxiv.org/abs/1709.01953}.

\bibitem[Neyshabur et~al.(2015)Neyshabur, Tomioka, and Srebro]{neyshabur2015}
Behnam Neyshabur, Ryota Tomioka, and Nathan Srebro.
\newblock In search of the real inductive bias: On the role of implicit
  regularization in deep learning, 2015.
\newblock URL \url{https://arxiv.org/abs/1412.6614}.

\bibitem[Phuong and Lampert(2020)]{Phuong2020}
Mary Phuong and Christoph~H. Lampert.
\newblock Functional vs. parametric equivalence of relu networks.
\newblock In \emph{International Conference on Learning Representations}, 2020.
\newblock URL \url{https://api.semanticscholar.org/CorpusID:214174225}.

\bibitem[Poggio et~al.(2018)Poggio, Liao, Miranda, Banburski, Boix, and
  Hidary]{Poggio2018}
Tomaso Poggio, Qianli Liao, Brando Miranda, Andrzej Banburski, Xavier Boix, and
  Jack Hidary.
\newblock Theory iiib: Generalization in deep networks, 2018.
\newblock URL \url{https://arxiv.org/abs/1806.11379}.

\bibitem[Rahimi and Recht(2007)]{Rahimi2007}
Ali Rahimi and Benjamin Recht.
\newblock Random features for large-scale kernel machines.
\newblock In J.~Platt, D.~Koller, Y.~Singer, and S.~Roweis, editors,
  \emph{Advances in Neural Information Processing Systems}, volume~20. Curran
  Associates, Inc., 2007.
\newblock URL
  \url{https://proceedings.neurips.cc/paper_files/paper/2007/file/013a006f03dbc5392effeb8f18fda755-Paper.pdf}.

\bibitem[Rochussen(2024)]{rochussen2024structuredpartialstochasticitybayesian}
Tommy Rochussen.
\newblock Structured partial stochasticity in bayesian neural networks, 2024.
\newblock URL \url{https://arxiv.org/abs/2405.17666}.

\bibitem[Rolnick and Kording(2020)]{rolnick2020a}
David Rolnick and Konrad Kording.
\newblock Reverse-engineering deep {R}e{LU} networks.
\newblock In Hal~Daumé III and Aarti Singh, editors, \emph{Proceedings of the
  37th International Conference on Machine Learning}, volume 119 of
  \emph{Proceedings of Machine Learning Research}, pages 8178--8187. PMLR,
  13--18 Jul 2020.
\newblock URL \url{https://proceedings.mlr.press/v119/rolnick20a.html}.

\bibitem[Rossi et~al.(2024)Rossi, Singh, and Hannagan]{Rossi2024}
Simone Rossi, Ankit Singh, and Thomas Hannagan.
\newblock On permutation symmetries in bayesian neural network posteriors: a
  variational perspective.
\newblock In \emph{Proceedings of the 37th International Conference on Neural
  Information Processing Systems}, NIPS '23, Red Hook, NY, USA, 2024. Curran
  Associates Inc.

\bibitem[Sargsyan et~al.(2015)Sargsyan, Najm, and Ghanem]{Sargsyan2015}
Khachik Sargsyan, Habib~N. Najm, and Roger Ghanem.
\newblock On the statistical calibration of physical models: Statistical
  calibration of physical models.
\newblock \emph{International Journal of Chemical Kinetics}, 47:\penalty0
  246--276, 2015.
\newblock URL \url{https://api.semanticscholar.org/CorpusID:93318877}.

\bibitem[Sargsyan et~al.(2018)Sargsyan, Huan, and Najm]{Sargsyan2018}
Khachik Sargsyan, Xun Huan, and Habib~N. Najm.
\newblock Embedded model error representation for bayesian model calibration.
\newblock \emph{International Journal for Uncertainty Quantification}, 2018.
\newblock URL \url{https://api.semanticscholar.org/CorpusID:86860281}.

\bibitem[Sharma et~al.(2023)Sharma, Farquhar, Nalisnick, and
  Rainforth]{pmlr-v206-sharma23a}
Mrinank Sharma, Sebastian Farquhar, Eric Nalisnick, and Tom Rainforth.
\newblock Do bayesian neural networks need to be fully stochastic?
\newblock In Francisco Ruiz, Jennifer Dy, and Jan-Willem van~de Meent, editors,
  \emph{Proceedings of The 26th International Conference on Artificial
  Intelligence and Statistics}, volume 206 of \emph{Proceedings of Machine
  Learning Research}, pages 7694--7722. PMLR, 25--27 Apr 2023.

\bibitem[Spiegelhalter and Lauritzen(1990)]{Spiegelhalter1990}
David~J. Spiegelhalter and Steffen~L. Lauritzen.
\newblock Sequential updating of conditional probabilities on directed
  graphical structures.
\newblock \emph{Networks}, 20\penalty0 (5):\penalty0 579--605, 1990.
\newblock \doi{https://doi.org/10.1002/net.3230200507}.
\newblock URL
  \url{https://onlinelibrary.wiley.com/doi/abs/10.1002/net.3230200507}.

\bibitem[Wang and Titterington(2005)]{wang05a}
Bo~Wang and D.~M. Titterington.
\newblock Inadequacy of interval estimates corresponding to variational
  bayesian approximations.
\newblock In Robert~G. Cowell and Zoubin Ghahramani, editors, \emph{Proceedings
  of the Tenth International Workshop on Artificial Intelligence and
  Statistics}, volume~R5 of \emph{Proceedings of Machine Learning Research},
  pages 373--380. PMLR, 06--08 Jan 2005.
\newblock URL \url{https://proceedings.mlr.press/r5/wang05a.html}.
\newblock Reissued by PMLR on 30 March 2021.

\bibitem[Wenzel et~al.(2020)Wenzel, Roth, Veeling, Swiatkowski, Tran, Mandt,
  Snoek, Salimans, Jenatton, and Nowozin]{pmlr-v119-wenzel20a}
Florian Wenzel, Kevin Roth, Bastiaan Veeling, Jakub Swiatkowski, Linh Tran,
  Stephan Mandt, Jasper Snoek, Tim Salimans, Rodolphe Jenatton, and Sebastian
  Nowozin.
\newblock How good is the {B}ayes posterior in deep neural networks really?
\newblock In Hal~Daumé III and Aarti Singh, editors, \emph{Proceedings of the
  37th International Conference on Machine Learning}, volume 119 of
  \emph{Proceedings of Machine Learning Research}, pages 10248--10259. PMLR,
  13--18 Jul 2020.
\newblock URL \url{https://proceedings.mlr.press/v119/wenzel20a.html}.

\bibitem[Zavatone-Veth et~al.(2022)Zavatone-Veth, Tong, and
  Pehlevan]{ZavatoneVeth2022ContrastingRA}
Jacob~A. Zavatone-Veth, William~L. Tong, and Cengiz Pehlevan.
\newblock Contrasting random and learned features in deep bayesian linear
  regression.
\newblock \emph{Physical review. E}, 105 6-1:\penalty0 064118, 2022.
\newblock URL \url{https://api.semanticscholar.org/CorpusID:247187786}.

\bibitem[Zhang et~al.(2017)Zhang, Bengio, Hardt, Recht, and Vinyals]{Zhang2017}
Chiyuan Zhang, Samy Bengio, Moritz Hardt, Benjamin Recht, and Oriol Vinyals.
\newblock Understanding deep learning requires rethinking generalization, 2017.
\newblock URL \url{https://arxiv.org/abs/1611.03530}.

\bibitem[Zhang et~al.(2021)Zhang, Bengio, Hardt, Recht, and Vinyals]{Zhang2021}
Chiyuan Zhang, Samy Bengio, Moritz Hardt, Benjamin Recht, and Oriol Vinyals.
\newblock Understanding deep learning (still) requires rethinking
  generalization.
\newblock \emph{Commun. ACM}, 64\penalty0 (3):\penalty0 107–115, feb 2021.
\newblock ISSN 0001-0782.
\newblock \doi{10.1145/3446776}.
\newblock URL \url{https://doi.org/10.1145/3446776}.

\end{thebibliography}


\clearpage
\newpage

\appendix

\section{Supplemental material}

\subsection{Extension to classification\label{A}}

We now consider an \(L\)-layer network in a classification setting:
\begin{eqnarray}
	\hat{y}  &=&  s_k (  w^\top x_L ), \\
	x_\ell  &=&  \sigma( \Theta_{\ell-1}^\top x_{\ell-1}  - b_{\ell-1} ), \qquad 1 < \ell \leq L, 
\end{eqnarray}
where \(s_k: \mathbb{R} \to [0,1]\).  The predictive class \(k\) belongs to \(\{0,1\}\) in the binary setting and \([K]\) in the multi-class case.  Let \(\Theta\) represent all interior parameters. Under the prior assumption, 
\begin{eqnarray}
	w \sim \Gauss ( 0, p^{-1}\mathbf{I}_p), \qquad
    \mathbb{P}  ( \Theta = \Theta^{(j)} )  = \rho_j, \qquad \sum_{j=1}^J \rho_j \ = \ 1,
\end{eqnarray}
we can write 
\begin{eqnarray}
	\label{eq:integral_w}
	\mathbb{P}(\tilde{y}=k \mid X_1, Y, \tilde{x}_1)  &=&  \sum_{j=1}^J \mathbb{P}( \Theta^{(j)} \mid X_1, Y) \; \mathbb{P}(\tilde{y}=k \mid X_1, Y, \tilde{x}_1, \Theta^{(j)}) \nonumber \\
	&=&  \sum_{j=1}^J \mathbb{P}( \Theta^{(j)} \mid X_1, Y) \; \int s_k(w^\top \tilde{x}_L) \; \pi( w \mid X_1, Y, \Theta^{(j)} ) \;  dw 
\end{eqnarray} 
where \(\mathbb{P}( \Theta^{(j)} \mid X_1, Y)\) is given by \eqref{eq:weight}. 

\subsubsection{Binary classification}

Suppose that \(y\in\{0,1\}\). Define \(s_1=s\) to be the logistic function. Then, the Gaussianity of \(\pi( w \mid X_1, Y, \Theta^{(j)} ) \) allows us to approximate the integral over \(w\), following \citep{MacKay1992a,Spiegelhalter1990,Kristiadi2020}. We use

\begin{eqnarray}
	s(\xi) \approx \Phi\left( \sqrt{\frac{\pi}{8}} \xi \right)
\end{eqnarray}
where \(\Phi\) is the probit function. We can now write
\begin{eqnarray}
	&& \mathbb{P}(\tilde{y}=1 \mid X_1, Y, \tilde{x}_1)  \\
	&\approx &  \sum_{j=1}^J \mathbb{P}( \Theta^{(j)} \mid X_1, Y)  \; \int\Phi\left( \sqrt{\frac{\pi}{8}} w^\top \tilde{x}_L \right) \; \pi( w^\top \tilde{x}_L  \mid X_1, Y, \tilde{x}_1, \Theta^{(j)} ) \; dw \nonumber \\[5pt]
	&=&  \sum_{j=1}^J \mathbb{P}( \Theta^{(j)} \mid X_1, Y)  \; \Phi\left( \frac{p^{-1}\widetilde{x}_L^\top X_L ( p^{-1}X_L^\top X_L \ + \ \gamma^2 \nonumber
	\mathbf{I} )^{-1} Y}{ \sqrt{ 8/\pi + \gamma^2   \ + \ \gamma^2 p^{-1} \widetilde{x}_L^\top  	
	( p^{-1}X_L X_L^\top \ + \ \gamma^2 \mathbf{I} )^{-1} \widetilde{x}_L  } } \right) \nonumber
\end{eqnarray}
Note that \(X_L\) and \(\tilde{x}_L\) depend on \(\Theta^{(j)}\).

\subsubsection{Multi-class classification}

Suppose that \(y\in\{1,\dots,K\}\). Then, \(w\in\mathbb{R}^{p \times K}\). Let \(s_k\) be the softmax function corresponding to class \(k\). We can then approximate the integral over \(w\) in \eqref{eq:integral_w} using the mean field approximation given by \citep{lu2021}:
\begin{eqnarray}
	&& \mathbb{P}(\tilde{y}=k \mid X_1, Y, \tilde{x}_1)  \\
	&=&  \sum_{j=1}^J \mathbb{P}( \Theta^{(j)} \mid X_1, Y)  \; \int s_k(  w^\top \tilde{x}_L ) \; \pi( w^\top \tilde{x}_L  \mid X_1, Y, \tilde{x}_1, \Theta^{(j)} ) \; dw \nonumber \\[5pt] 
	&=& \sum_{j=1}^J \mathbb{P}( \Theta^{(j)} \mid X_1, Y)  \; \int \left(  2 - K + \sum_{r \neq k} \frac{1}{\texttt{sigmoid}( w^{(k)\top} \tilde{x}_L - w^{(r)\top} \tilde{x}_L )} \right)^{-1} \times \nonumber \\[6pt] && \qquad \qquad \qquad \qquad \qquad \qquad  \pi( w^\top \tilde{x}_L  \mid X_1, Y, \tilde{x}_1, \Theta^{(j)} ) \; dw  \nonumber\\[5pt] 
	&\approxtext{mean field}& \sum_{j=1}^J \mathbb{P}( \Theta^{(j)} \mid X_1, Y)  \; \left(  2 - K + \sum_{r \neq k} \frac{1}{  \mathbb{E}_{\pi(k,r)}[ \texttt{sigmoid}(  w^{(k)\top} \tilde{x}_L - w^{(r)\top} \tilde{x}_L) ] } \right)^{-1} \nonumber
\end{eqnarray}
where \(w^{(r)}\) is the \(r^{th}\) column of \(w\) and \(\pi(k,r)\) is the joint distribution of \( w^{(k)\top} \tilde{x}_L\) and \( w^{(r)\top} \tilde{x}_L\) conditioned on \(X_1, Y, \) and \(\Theta^{(j)}\), which is obtained based on \eqref{eq:y_dist}.

\subsection{PDFs under a discretized Gaussian prior \label{B}}

\begin{figure*}
	\includegraphics[width=0.95\linewidth]{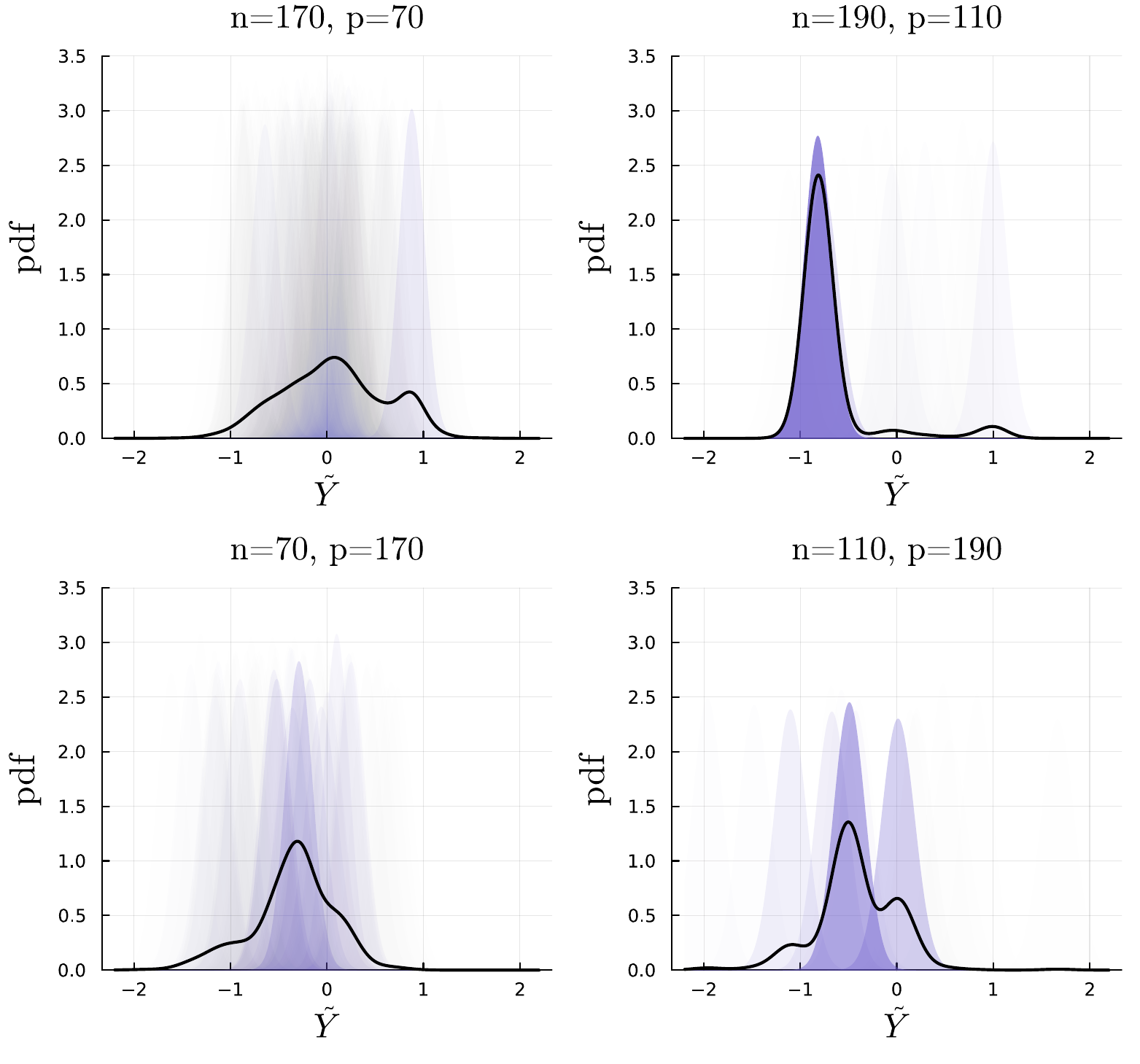} 
    \caption{\label{fig:empirical_pdfs2}  Posterior predictive distributions at test point \(\widetilde{x}_1^{(2)}\) for input dimension \(d=100\) at select training set sizes \(n\) and final layer widths \(p\), as indicated by each title. The black line shows the pdf which is a mixture of Gaussians. Each shaded distribution is a component of this mixture with transparency corresponding to its weight.  }
\end{figure*}

\begin{figure*}
	\includegraphics[width=0.95\linewidth]{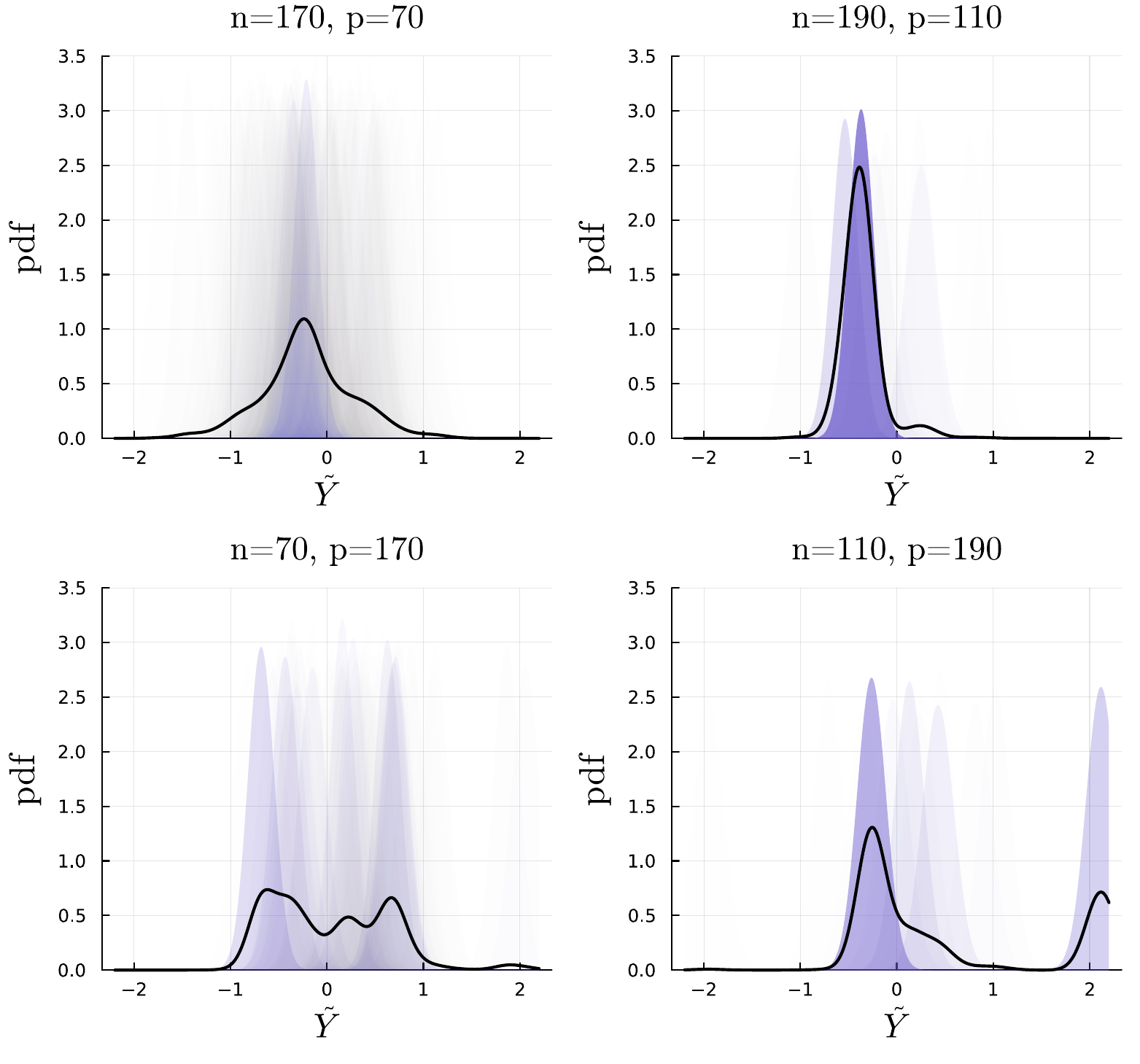} 
    \caption{\label{fig:empirical_pdfs3} Posterior predictive distributions at test point \(\widetilde{x}_1^{(3)}\) for input dimension \(d=100\) at select training set sizes \(n\) and final layer widths \(p\), as indicated by each title. The black line shows the pdf which is a mixture of Gaussians. Each shaded distribution is a component of this mixture with transparency corresponding to its weight. }
\end{figure*}

\begin{figure*}
	\includegraphics[width=0.95\linewidth]{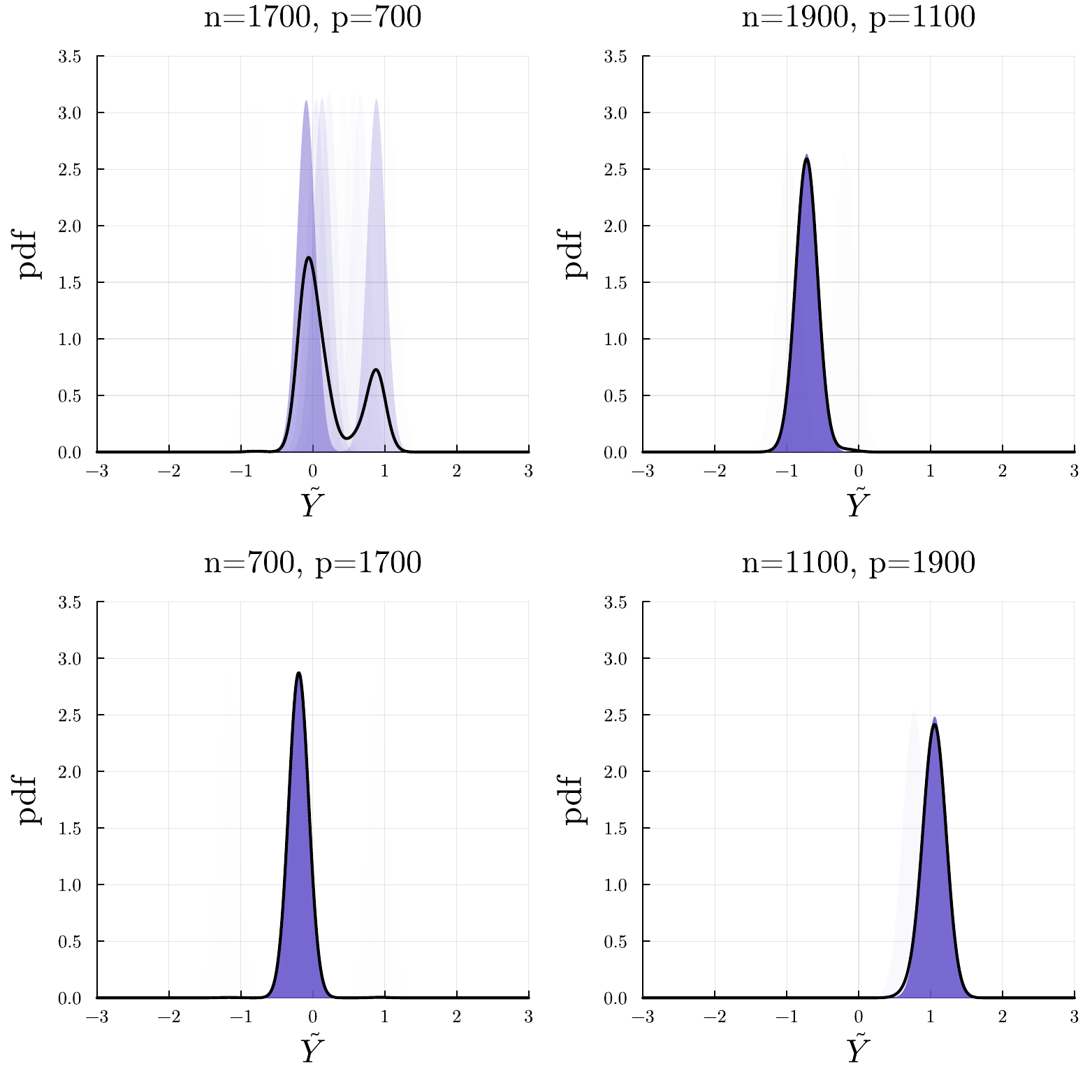} 
    \caption{\label{fig:empirical_pdfs2_1000} Posterior predictive distributions at test point \(\widetilde{x}_1^{(2)}\) for input dimension \(d=1000\) at select training set sizes \(n\) and final layer widths \(p\), as indicated by each title. The black line shows the pdf which is a mixture of Gaussians. Each shaded distribution is a component of this mixture with transparency corresponding to its weight.  }
\end{figure*}

\begin{figure*}
	\includegraphics[width=0.95\linewidth]{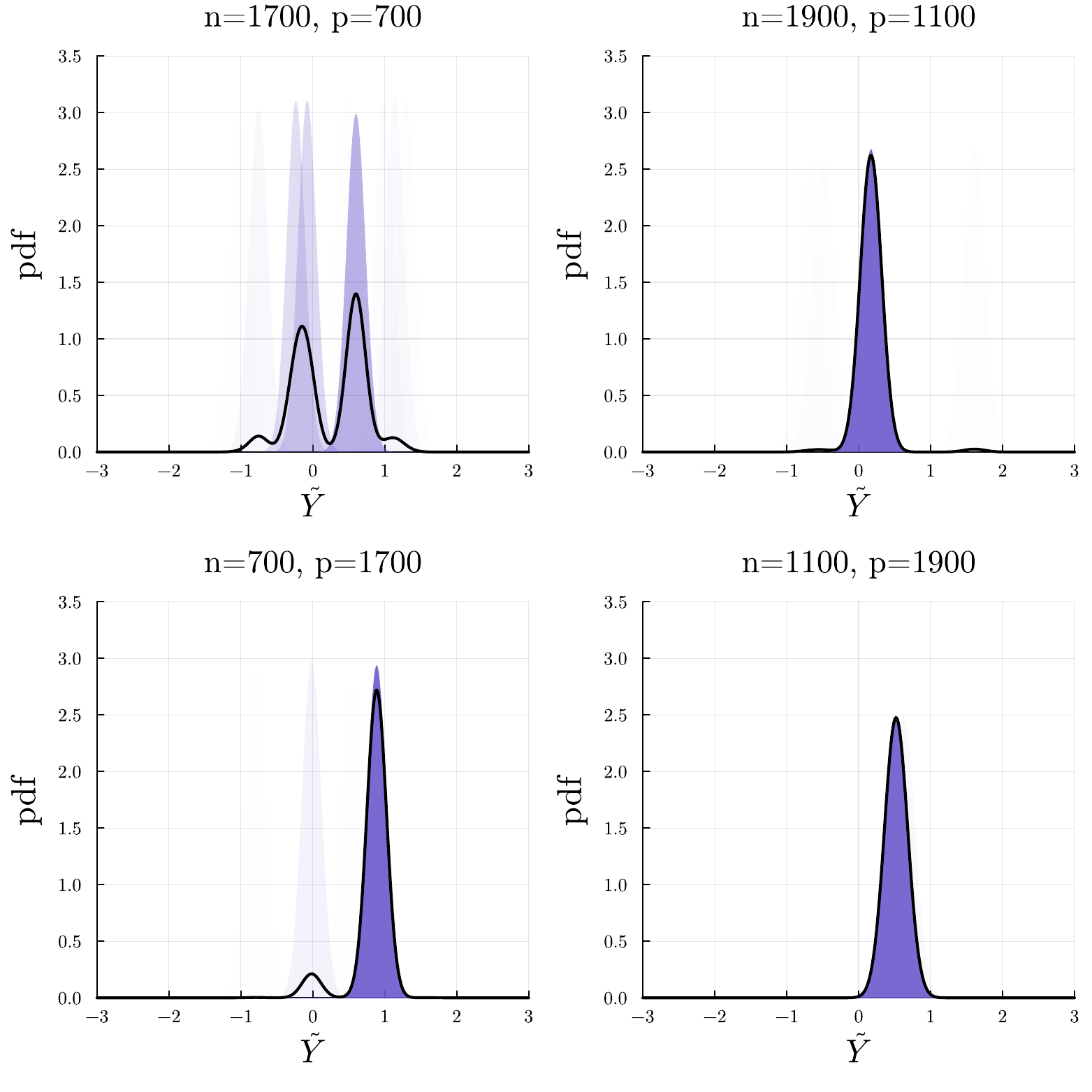} 
    \caption{\label{fig:empirical_pdfs3_1000} Posterior predictive distributions at test point \(\widetilde{x}_1^{(3)}\) for input dimension \(d=1000\) at select training set sizes \(n\) and final layer widths \(p\), as indicated by each title. The black line shows the pdf which is a mixture of Gaussians. Each shaded distribution is a component of this mixture with transparency corresponding to its weight. }
\end{figure*}

Figure \ref{fig:empirical_pdfs} shows the predictive distributions for select network and training set sizes at test location \(\widetilde{x}_1^{(1)}\). Here, we provide the pdfs which result from inference with the prior specified in Section \ref{gaussian} for the same ratios \(n/p\) at additional locations \(\widetilde{x}_1^{(2)}\) and \(\widetilde{x}_1^{(3)}\). For all examples, \(\gamma^2 = 0.01\). Figures \ref{fig:empirical_pdfs2} and \ref{fig:empirical_pdfs3} show results for \(d=100\) while figures \ref{fig:empirical_pdfs2_1000} and \ref{fig:empirical_pdfs3_1000} correspond to \(d=1000\). As in Section \ref{gaussian}, we see that multiple candidates \(\Theta^{(j)}\) contribute to the posterior predictive distributions, leading to multimodality in most cases considered. We observe fewer cases of multimodality for larger \(d\) due to the double descent effect in inference for \(w\) shown in the heatmaps of Figure \ref{fig:empirical_pdfs}. As we found in Section \ref{constructed}, unimodal predictive distributions for \(p\) close to \(n\) are an artifact of choosing parameter candidates to have elements independently sampled from Gaussian distributions.

\clearpage
\newpage

\subsection{Identifying modes under a discretized Gaussian prior \label{C}}

\begin{figure*}
	\includegraphics[width=0.95\linewidth]{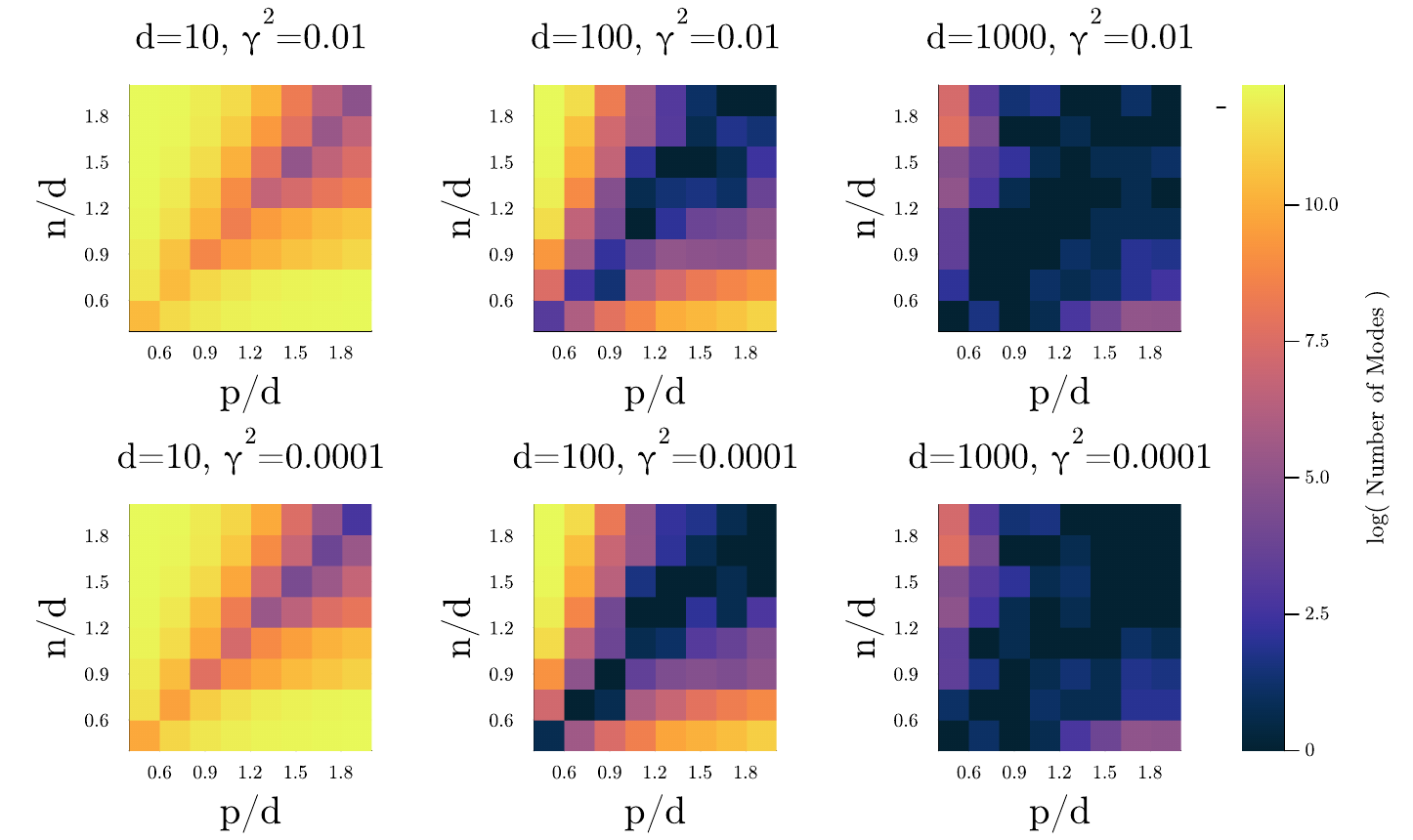} 
    \caption{\label{fig:nmodes}   Heatmaps depicting the log of the number of component distributions which have weight larger than \(10^{-6}\) for specified network dimensions.  Columns correspond to the input dimension, \(d\): \(10\), \(100\), and \(1000\). Rows correspond to observation noise variance: \(0.01\) and \(0.0001\).    }
\end{figure*}

\begin{figure*}
	\includegraphics[width=0.95\linewidth]{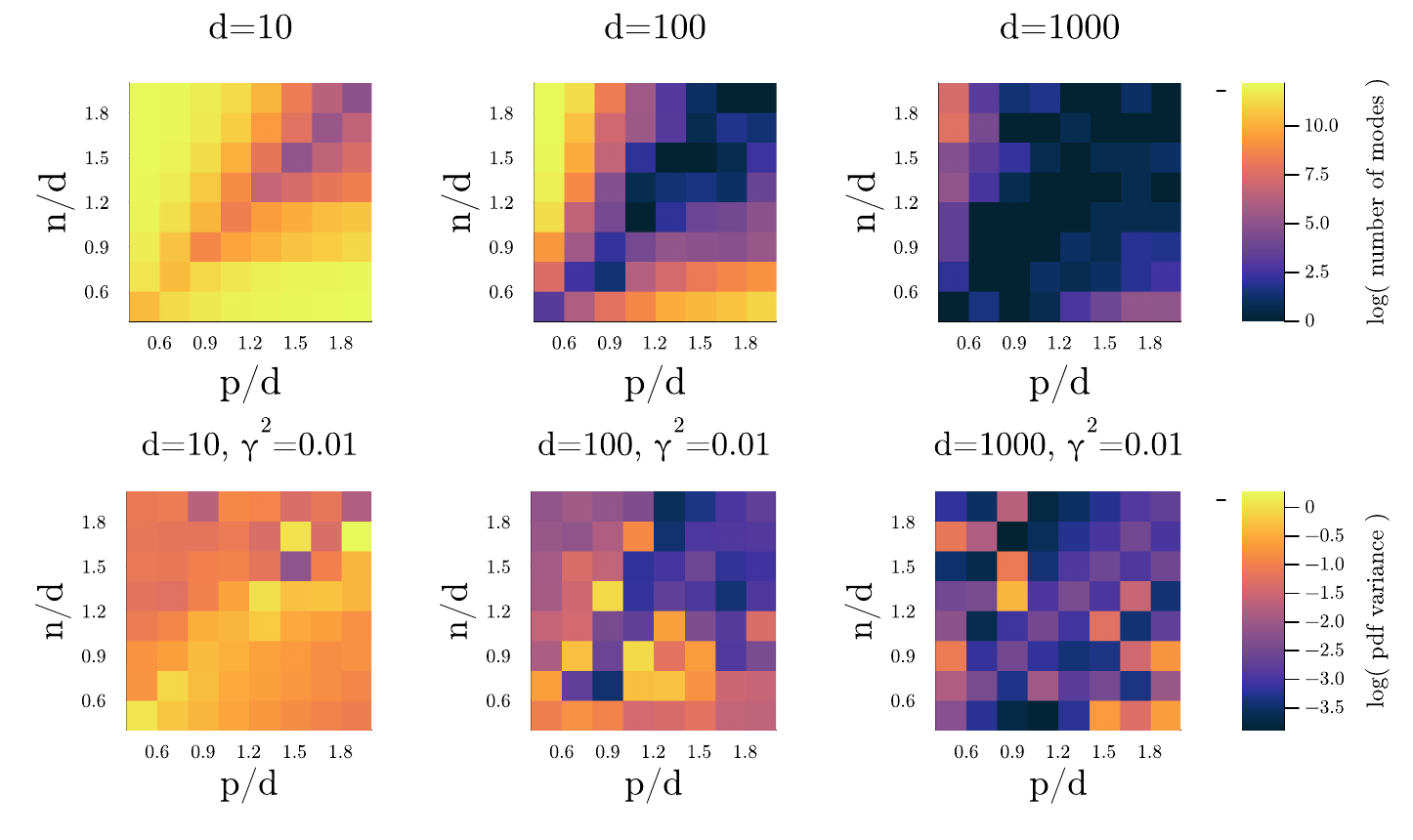} 
    \caption{\label{fig:nmodes_var} Top: The log of the number of component distributions which have weight larger than \(10^{-6}\) for specified network dimensions.  Bottom: The log of the variance of the posterior predictive distribution obtained for each network size. Columns correspond to the input dimension, \(d\): \(10\), \(100\), and \(1000\). All results correspond to observation noise \(\gamma^2=0.01\).  }
\end{figure*}

\begin{figure*}
	\includegraphics[width=0.7\linewidth]{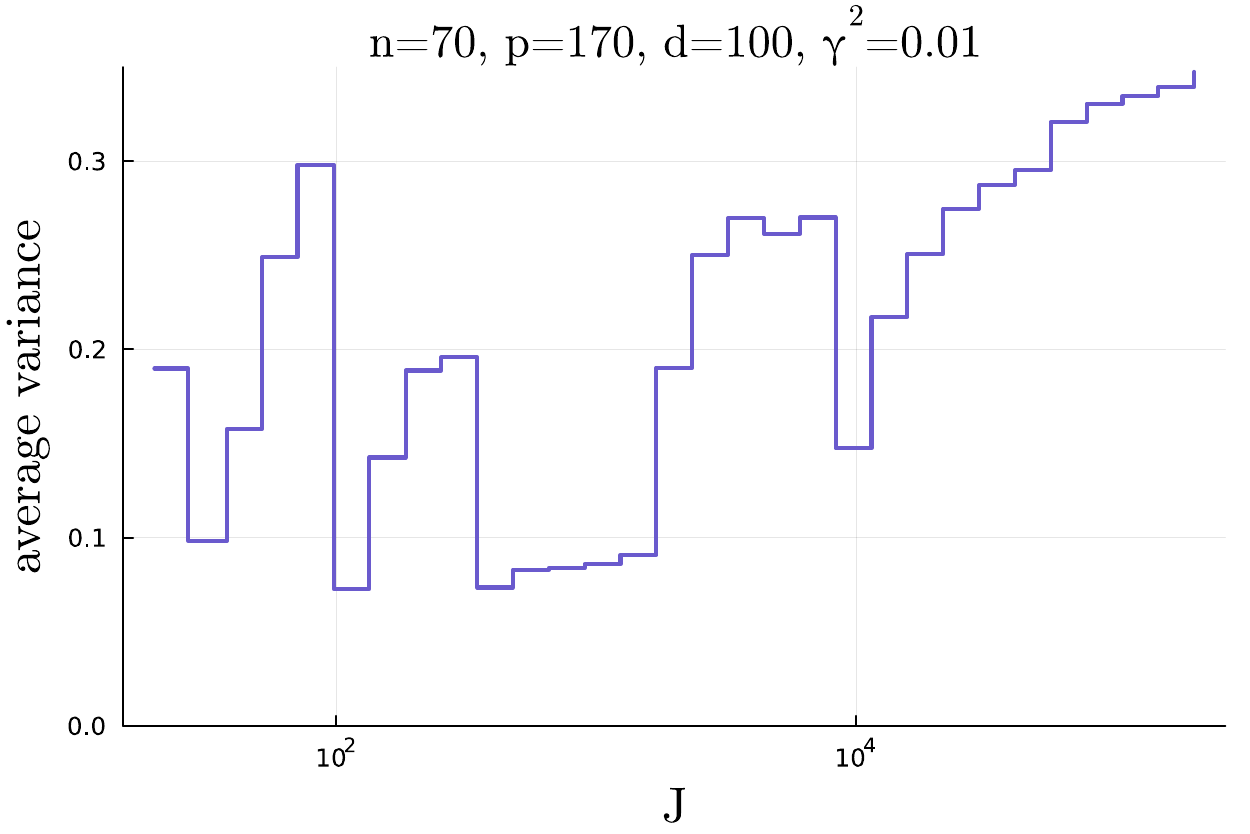} 
    \caption{\label{fig:jsensitivity}  The variance of the posterior predictive distribution averaged over \(100\) realizations of \(\widetilde{x}_1\) plotted against the number of parameter candidates, \(J\). The prior details are specified in Section \ref{gaussian} and the network and training set size are given in the plot title.   }
\end{figure*}

\begin{figure*}
	\includegraphics[width=0.9\linewidth]{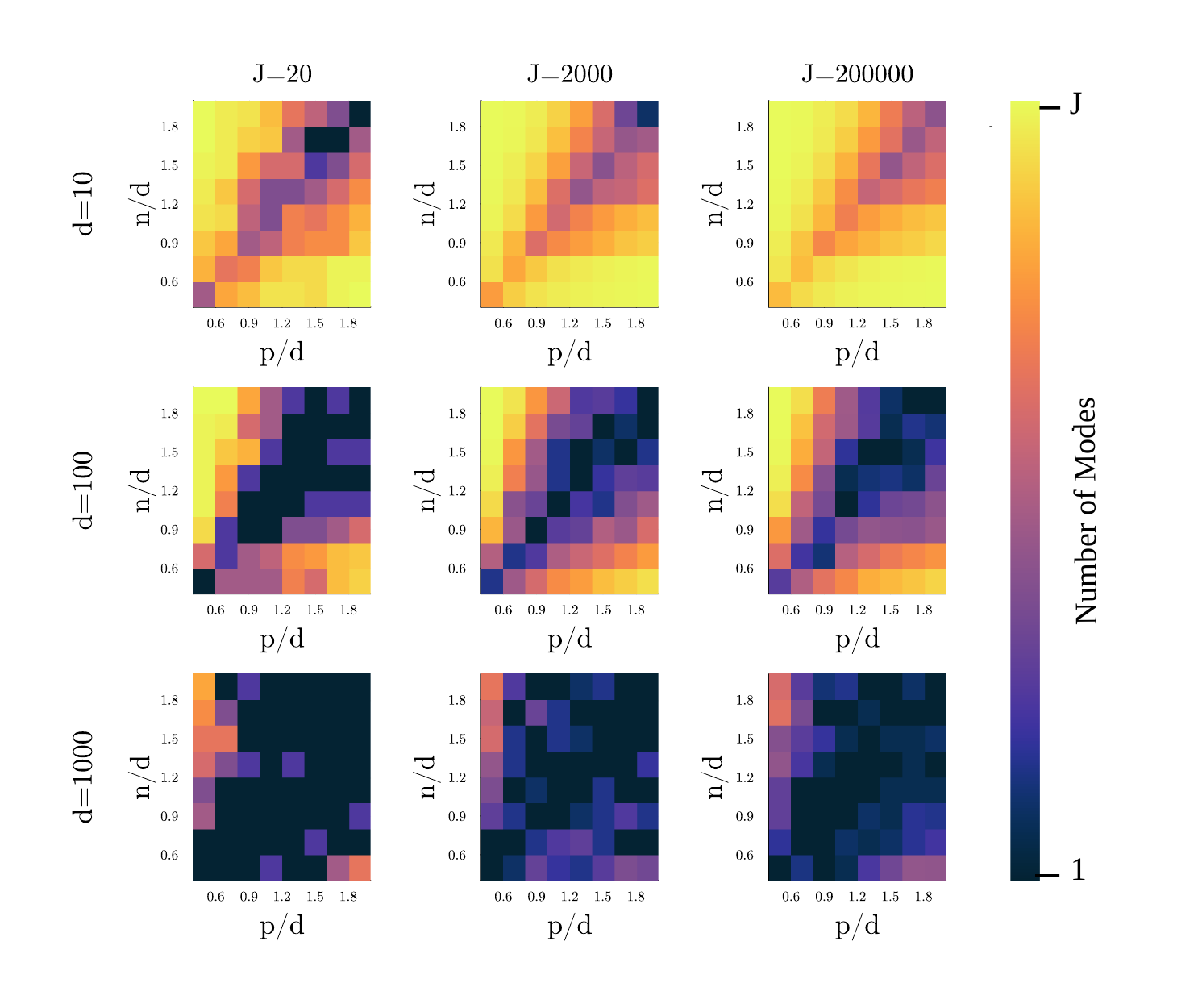} 
    \caption{\label{fig:jcomp} A comparison of the number of significant modes in the posterior predictive distribution for various network sizes and \( J \in \{20, 2000, 200\,000\} \) for the setting of Section \ref{gaussian}. Rows correspond to different covariate dimensions, \(d\), and columns correspond to different parameter candidate set sizes, \(J\).   Observation noise variance is set to \(0.01\).}
\end{figure*}

This section provides additional results concerning the multimodality and variance of predictive distributions described in Section \ref{gaussian}. The first rows of both Figures \ref{fig:nmodes} and \ref{fig:nmodes_var} match the right column of Figure \ref{fig:empirical_pdfs}. This set of heatmaps reports the number of modes with weight larger than \(10^{-6}\) found for specified network and training set size at observation noise level \(\gamma^2=0.01\). They are repeated for the purpose of comparison. In Figure \ref{fig:nmodes}, we see that the number of modes located for a specific \(n,p,d\) triple is not impacted by reducing observation noise to \(\gamma^2=0.0001\). In Figure \ref{fig:nmodes_var}, we can see a loose relationship between the number of significant modes and the variance of the predictive distribution. In our numerical experiments, we have found that component distributions of the predictive distribution tend to have similar variance and may have distinct modes. Thus, it is reasonable that finding more significant modes correlates with greater predictive variance, as we see. For this particular example, as we increase the training set size, predictive variance tends to decrease, but this may be an artifact of the prior choice and finite \(J\). Figure \ref{fig:jsensitivity} demonstrates that the predictive variance can be sensitive to the choice of \(J\). Note, however, that our findings on multimodality for various network sizes do not require \(J\) to be large. Figure \ref{fig:jcomp} shows that similar patterns are seen for \( J\in\{20, 2000, 200\,000\}\) at the considered values of \(d\), \(n/d\), and \(p/d\).

\begin{figure*}
	\includegraphics[width=0.95\linewidth]{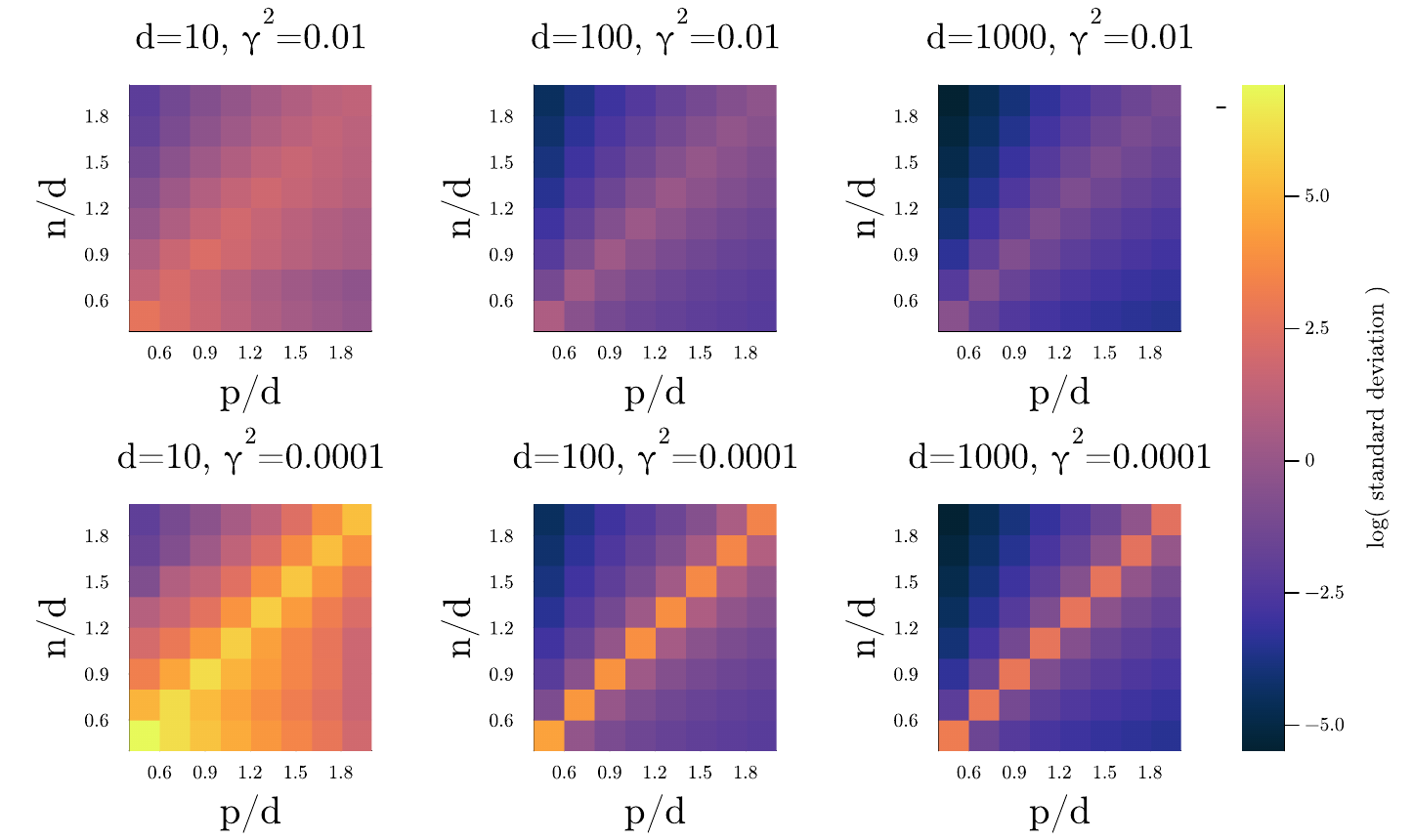} 
    \caption{\label{fig:stdev}  Heatmaps depicting the log of the standard deviation of the distribution of \(n^{-1}\log\mathcal{L}(\Theta)\) for candidates \(\Theta^{(j)}\) sampled from a Gaussian prior as described in Section \ref{gaussian}. Columns correspond to the input dimension, \(d\): \(10\), \(100\), and \(1000\). Rows correspond to observation noise variance: \(0.01\) and \(0.0001\). The diagonal where \(n=p\) shows a double descent effect which is stronger for smaller observation noise.}
\end{figure*}

Figure \ref{fig:stdev} provides context for the number of modes reported in Figure \ref{fig:nmodes}. The heatmap shades correspond to the log of the standard deviation of the distribution on  \(\{ n^{-1}\log\mathcal{L}(\Theta^{(j)}) \}_{j=1}^J\).  For a given input dimension, \(d\), and observation noise level, \(\gamma^2\), the largest standard deviation is found when \(n=p\). This effect is likely related to the double descent phenomena: for each \(X_1\), there is one candidate \(\Theta\) which outperforms all other candidates. As expected, the double descent phenomenon becomes more pronounced as regularization, \(\gamma^2\), decreases.

\clearpage
\newpage

\subsection{Optimal parameters\label{D}}

We are interested in \(\Theta\) which maximizes \(\mathcal{L}(\Theta;X,Y)\) as defined in \eqref{eq:likelihood}. To this end, consider the singular value decomposition
\begin{eqnarray}
	U \Lambda^{1/2} Q^\top \ = \ \frac{X_L}{\sqrt{p}}, 
\end{eqnarray}
where \(\mathrm{diag}(\Lambda)=[\lambda_1,\dots,\lambda_{p\vee n}]^\top\) and \(Q=[q_1\dots q_n]\ \in\mathbb{R}^{n\times n}\). Then, 
\begin{eqnarray}
	\log\mathcal{L}\left( X_L(\Theta) ; X,Y\right) = \frac{-1}{2} \sum_{k=1}^n\left(  \log(2\pi) + \log(\lambda_k +\gamma^2) + \frac{(q_k^\top Y)}{\lambda_k+\gamma^2}\right).
\end{eqnarray} 
Note that because \(X_L^\top X_L\) is positive semi-definite, \(n^{-1} \log\mathcal{L}\left( X_L \right) \ \leq \ - \log(\gamma)\). We can determine that
\begin{eqnarray*}
	&& \min_\Theta  \frac{-2}{n}\left( \log\mathcal{L}\left( X_L(\Theta) ; X,Y \right) + \log(2\pi) \right) \\ 
	&\leq &   \min_{\substack{\Lambda \succeq 0 \\ Q^\top Q = QQ^\top = \mathbf{I}_n}} \ \frac{1}{n}\sum_{k=1}^n\left(  \log(\lambda_k +\gamma^2) + \frac{(q_k^\top Y)}{\lambda_k+\gamma^2}\right) \\	
	&=&   \min_{ Q^\top Q = QQ^\top = \mathbf{I}_n} \ \frac{1}{n}\sum_{k=1}^n  \min_{\lambda_k \geq 0}\left(  \log(\lambda_k +\gamma^2) + \frac{(q_k^\top Y)}{\lambda_k+\gamma^2}\right)  \\
	&=& \  \min_{  Q^\top Q = QQ^\top = \mathbf{I}_n }    \frac{1}{n}\sum_{k=1}^n  
    \begin{cases}
        \log\left( q_k^\top Y\right)^2 + 1 & \quad   (q_k^\top Y)^2 \geq \gamma^2, \ k \leq p \\[5pt]
        \log\gamma^2 + \frac{(q_k^\top Y)}{\gamma^2} & \quad \mathrm{otherwise}
    \end{cases} \\[7pt]
    &=& \ \log\gamma^2 \ + \      \min_{\substack{ \{v_1\geq\dots\geq v_n\geq 0,\\[3pt] \gamma^2\sum_{i=1}^n v_i = Y^\top Y \} } }    \frac{1}{n}\sum_{k=1}^n  
    \begin{cases}
        \log v_k + 1 & \quad   v_k \geq 1, \ k \leq p \\[5pt]
        v_k & \quad \mathrm{otherwise}
    \end{cases}.
\end{eqnarray*} 
In the last line, we impose the constraint \(v_1\geq\dots\geq v_n\geq 0\) to prevent redundant optima. We find that
\begin{eqnarray}
	\argminA_{X_L^\top X_L}  \frac{1}{n}\sum_{k=1}^n\left(  \log(\lambda_k +\gamma^2) + \frac{(q_k^\top Y)}{\lambda_k+\gamma^2}\right) \ = \ YY^\top\left(1-\frac{\gamma^2}{Y^TY}. \right)
\end{eqnarray}
Provided that \(Y^\top Y\geq\gamma^2\), this minimizer is unique. For the results reported in this work, we assume that \(Y\in\mathbb{R}^n\) is centered with unit variance. Then, we expect \(Y^\top Y \sim \mathcal{O}(n)\). 

\clearpage
\newpage

\subsection{Optimal parameters for ReLU \label{E}}

In this section, we evaluate the conjecture made by \eqref{eq:conjecture}. In particular, we compare the largest value of \(n^{-1}\log\mathcal{L}(X_L)\) found in Section \ref{gaussian} to the conjectured maximum for a given set of training observations \((X_1,Y)\). Recall that for a two-layer network, we must have \(n\leq d\) for there to exist some \(\Theta\) which maps to the conjectured maximizer, \(X_L^*\). Thus, for this section we consider \( \mathcal{L}\left(X_L(\Theta)\right) \) such that  
\begin{eqnarray}
	\label{eq:project_conjecture}
	X_L^\top X_L \ =  \ \sigma(P_X Y Y^\top P_X^\top )\left(1 - \frac{\gamma^2}{Y^\top Y}  \right)
\end{eqnarray}
where \(P_X\) projects into the column space of \(X_1\). Thus, when \(n\leq d\), \eqref{eq:project_conjecture} reduces to \eqref{eq:conjecture}. Note that we do not necessarily expect  \(n^{-1}\log\mathcal{L}(X_L)\) under \eqref{eq:project_conjecture} to be optimal when \(n>d\). 

\begin{figure*}
	\includegraphics[width=0.95\linewidth]{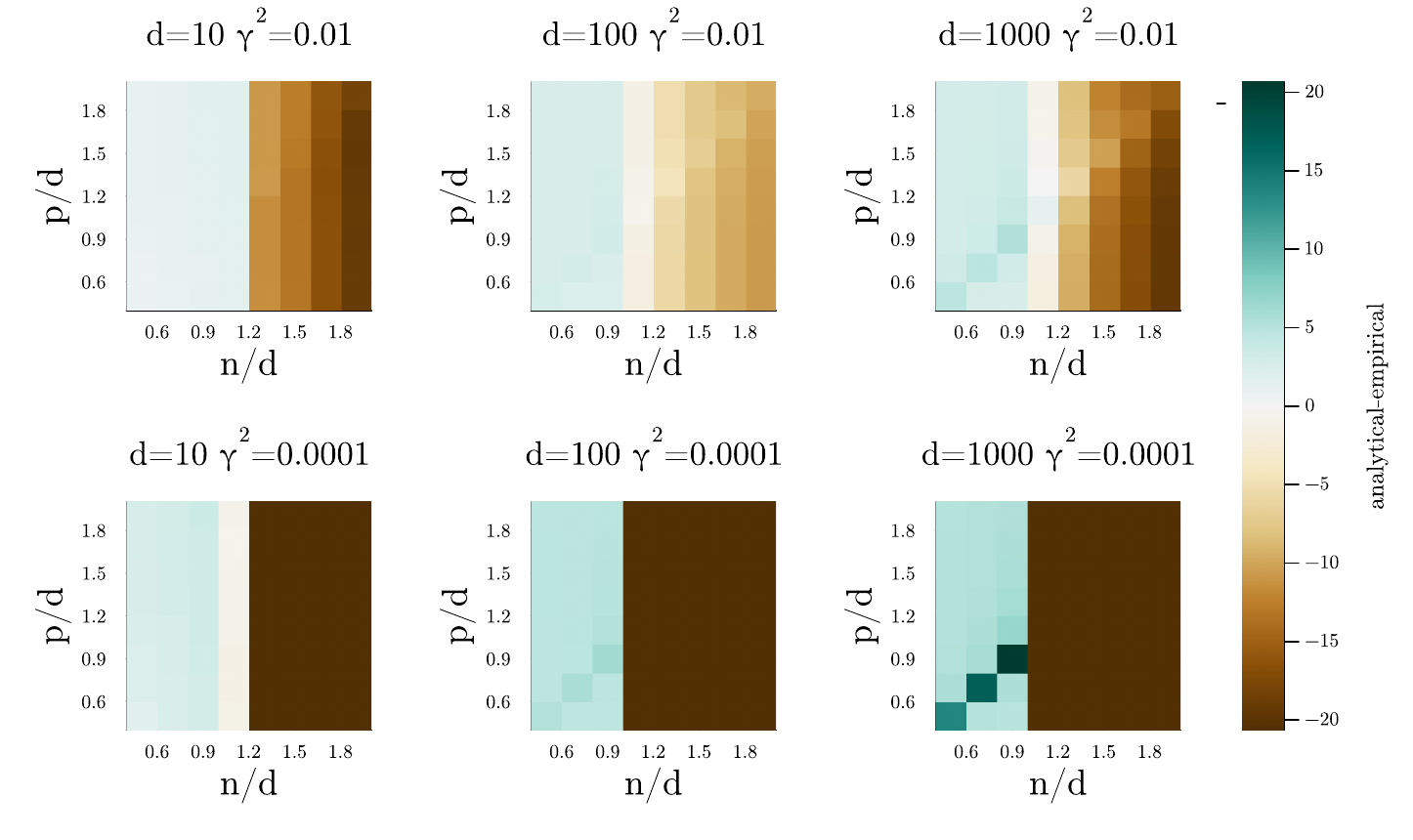} 
    \caption{\label{fig:bounds} The difference in scaled log marginal likelihood (\(n^{-1}\log\mathcal{L}\)) based on \(\Theta\) constructed to satisfy \eqref{eq:project_conjecture} and the best performing \(\Theta\) with elements sampled iid from a Gaussian prior. }
\end{figure*}

Figure \ref{fig:bounds} shows the difference between optimal \(n^{-1}\log\mathcal{L}(X_L)\) under \eqref{eq:project_conjecture} and the maximum \(n^{-1}\log\mathcal{L}(X_L)\) found empirically in Section \ref{gaussian}. We consider \(d\in\{10,100,1000\}\), \(\gamma^2\in\{0.01,0.0001\}\), and ratios \(p/d\) and \(n/d\) ranging from \(0.5\) to \(2\).  As expected, we find that the conjectured optimum is at least as large as the empirically determined maximum for \(n\leq d\). In cases where \(d<n\), enforcing \eqref{eq:project_conjecture} leads to \(n^{-1}\log\mathcal{L}(X_L)\) considerably smaller than our empirically located maximum. It is interesting to note that the distance by which the conjectured maximum outperforms the empirical maximum is exacerbated when \(d\) is large and \(n=p\).

\clearpage
\newpage

\subsection{Predictive distribution for optimal parameters \label{F}}

\begin{figure*}
	\begin{tabular}{cc}
	\includegraphics[width=0.45\linewidth]{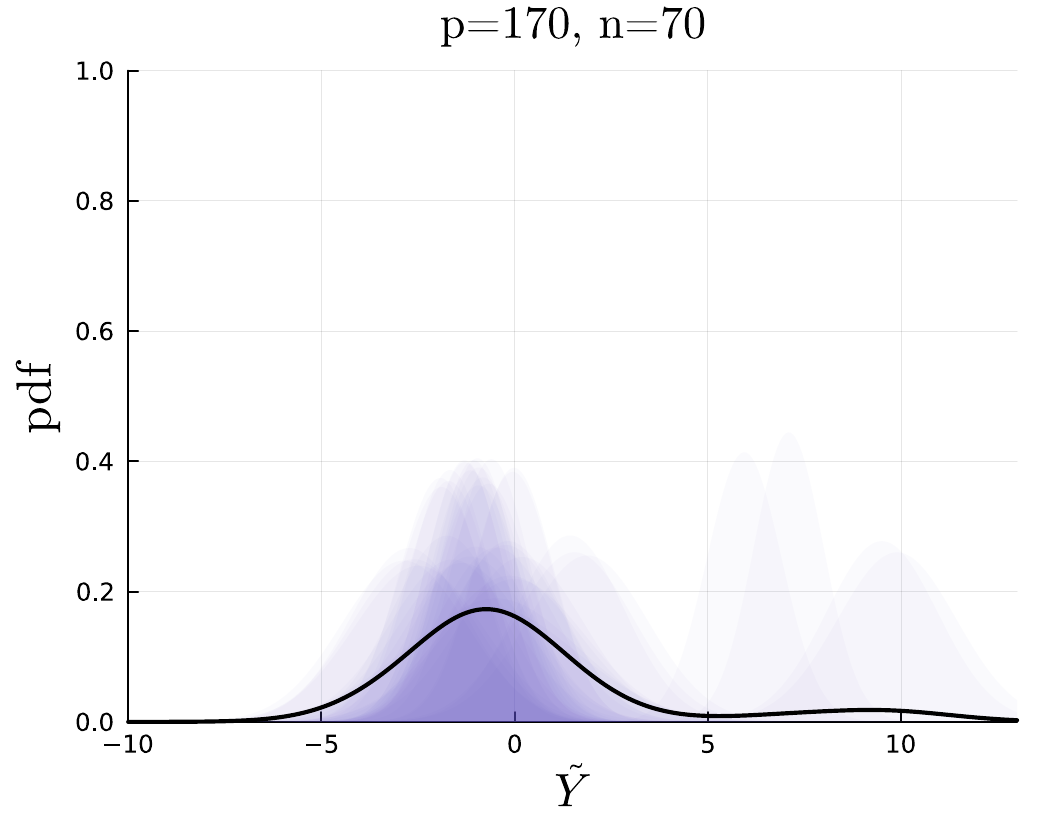} & 
	\includegraphics[width=0.45\linewidth]{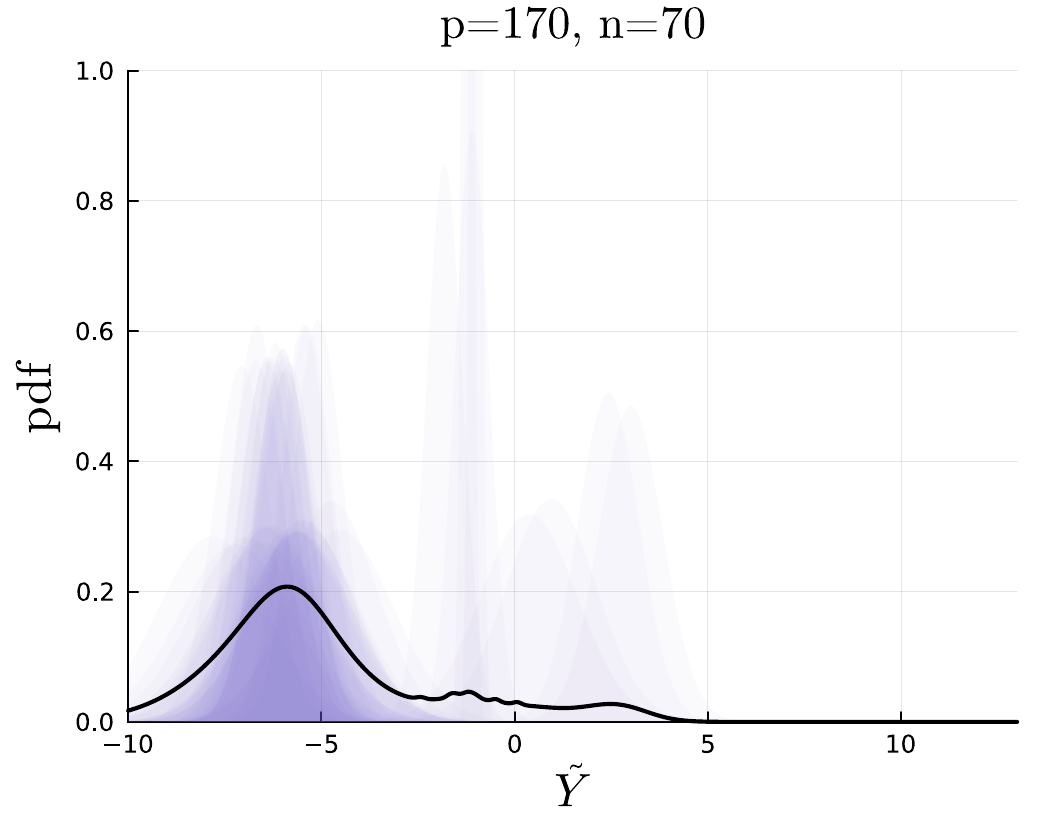}
	\end{tabular}
    \caption{\label{fig:opt_pdf23} Predictive distribution based on candidate parameters constructed to achieve \eqref{eq:conjecture}. The full distribution is plotted in black and components are shaded according to their weight in indigo. We consider \(10\) rotations, \(10\) preimage samples, and \(10\) column space samples to construct the distribution. Results are for \(n=70\), \(p=170\), and \(d=100\) at locations \(\widetilde{x}_1^{(2)}\) (left) and \(\widetilde{x}_1^{(3)}\) (right).    }
\end{figure*}

\begin{figure*}
	\begin{tabular}{c}
	\includegraphics[width=0.95\linewidth]{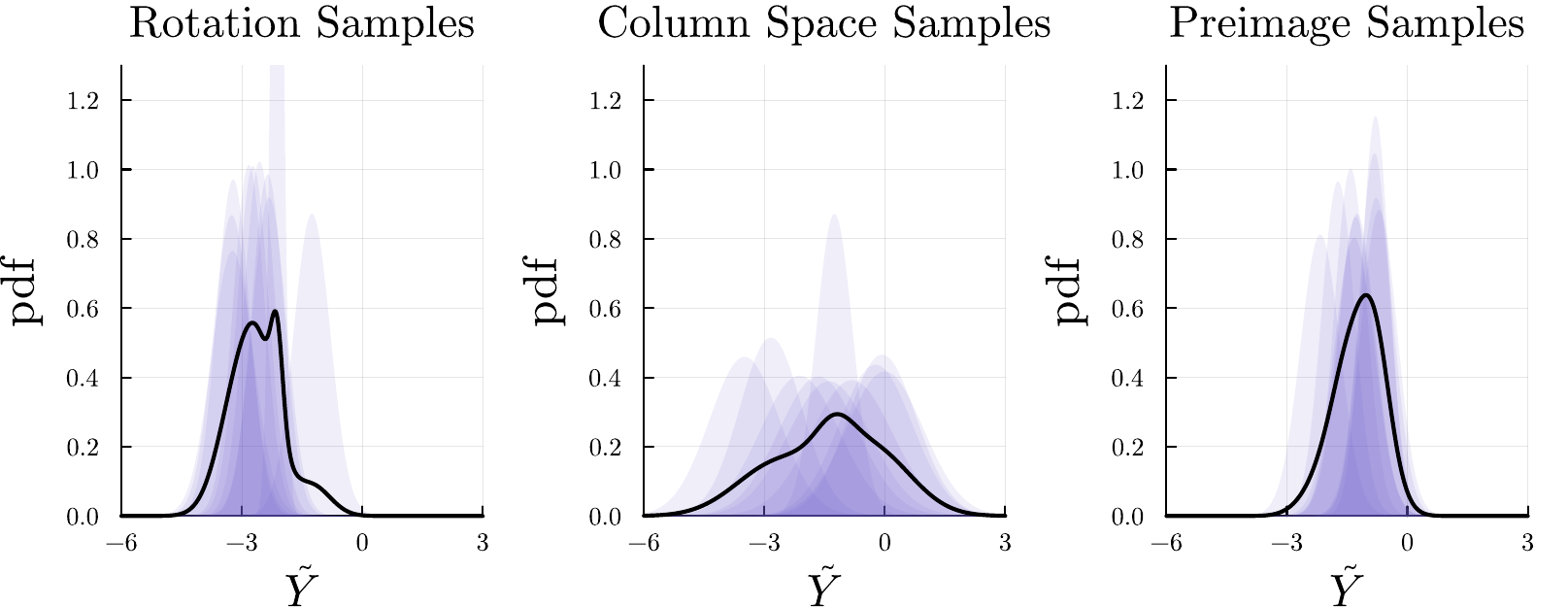} \\
	\includegraphics[width=0.95\linewidth]{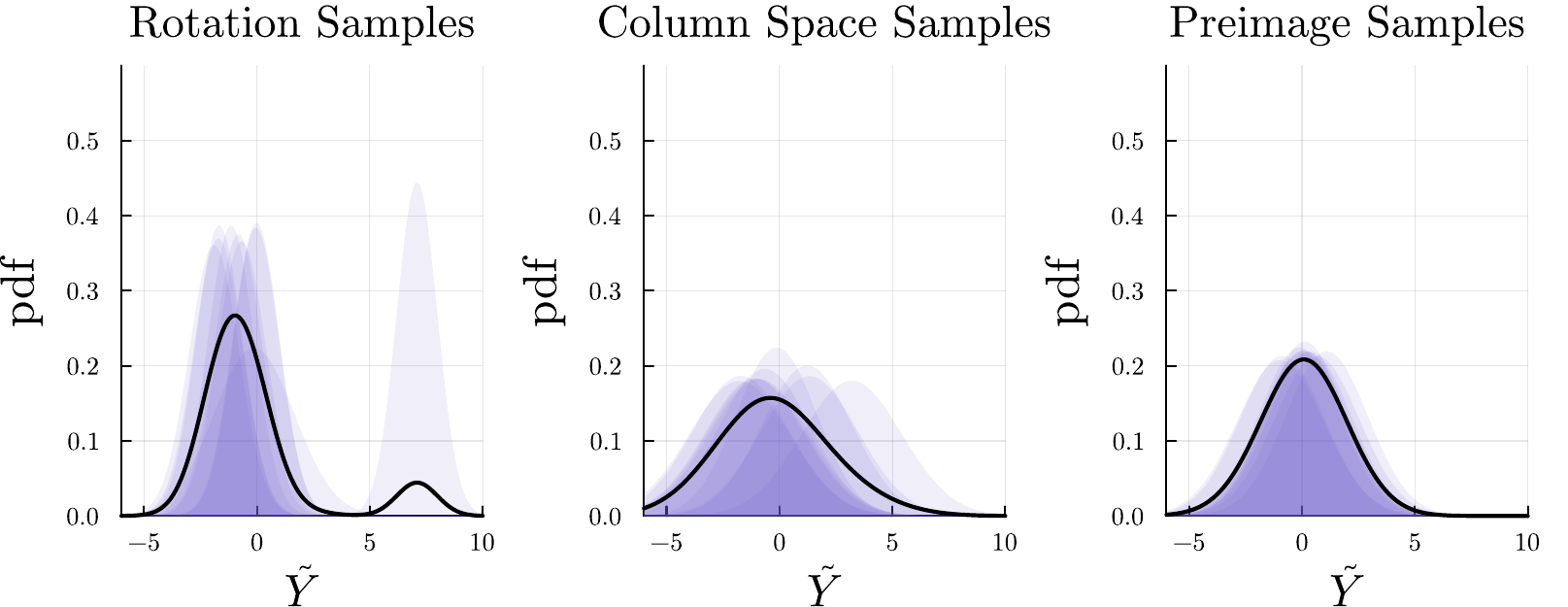} \\
	\includegraphics[width=0.95\linewidth]{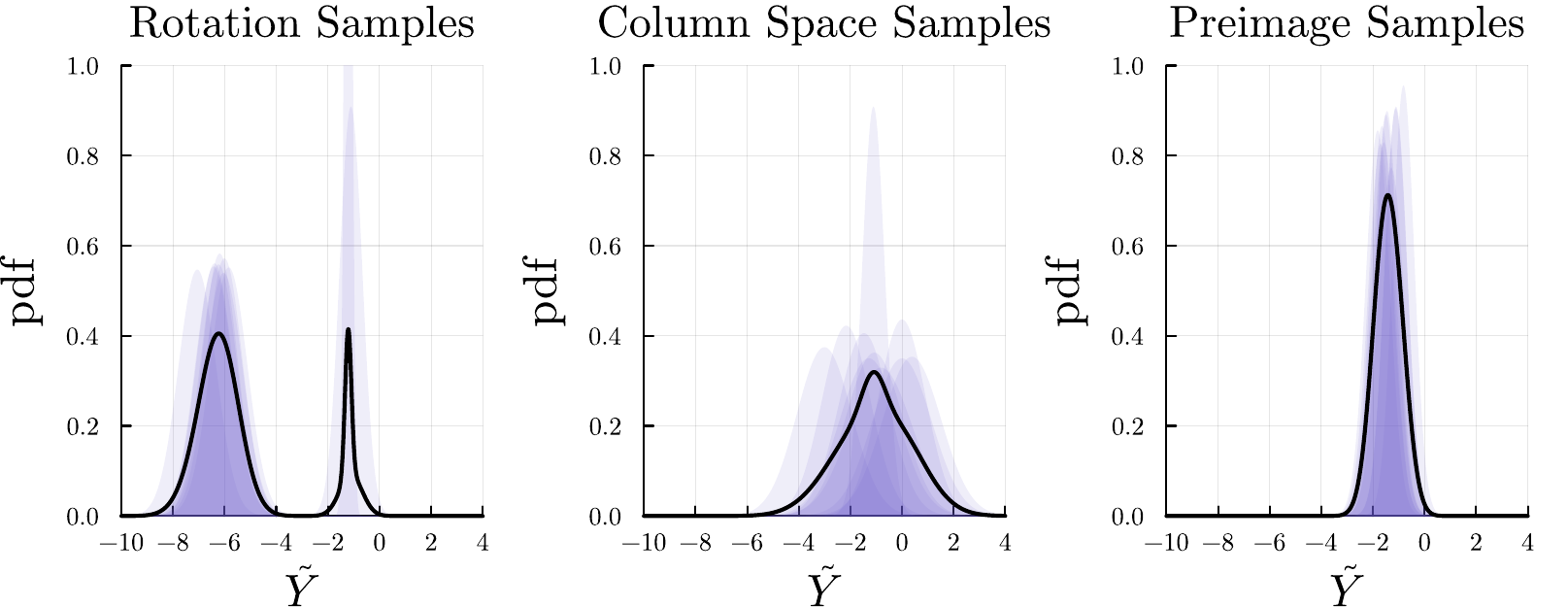}
	\end{tabular}
    \caption{\label{fig:opt_pdf_separated} Predictive distributions based on \(10\) parameter candidates differentiated by the method of their construction: rotation (left), sampling the column space (center), sampling the preimage (right). For all results, \(n=70\), \(p=170\), and \(d=100\). Each row corresponds to a different test location:  \(\widetilde{x}_1^{(1)}\), \(\widetilde{x}_1^{(2)}\), and \(\widetilde{x}_1^{(3)}\).  }
\end{figure*}

Here, we summarize additional results from the setting of Section \ref{constructed}. The top left subfigure of Figure \ref{fig:optimized_pdfs} shows the predictive distributions for \(n=70\) and \(p=170\) at test location \(\widetilde{x}_1^{(1)}\). Here, we provide the predictive distribution at additional locations \(\widetilde{x}_1^{(2)}\) and \(\widetilde{x}_1^{(3)}\) in Figure \ref{fig:opt_pdf23}. For all examples, \(\gamma^2 = 0.01\). The candidates \(\Theta^{(j)}\) for these plots are constructed from the combination of \(10\) rotations of \(X_L\), \(10\) samples of the column space of \(X_1\), and \(10\) samples of the preimage space. Thus, we have a total of \(1000\) candidates. To better separate the impact of each approach to constructing candidates, Figure \ref{fig:opt_pdf_separated} shows predictive pdfs where each column corresponds to a different approach. For instance, in the first column, candidates are constructed based on  \(10\) rotations of \(X_L\), one sample of the column space of \(X_1\), and one sample of the preimage space. Each row corresponds to a different test location: \(\widetilde{x}_1^{(1)}\), \(\widetilde{x}_1^{(2)}\) and \(\widetilde{x}_1^{(3)}\). We see that all approaches see to contribute to predictive variance, but rotation and column space samples seem to yield more distinct modes than preimage samples. 

Figure \ref{fig:nmodes_var} suggests that under the setting of Section \ref{gaussian}, as \(n\) increases, the variance of the predictive distribution decreases, even if \(p\) and \(d\) increase in proportion to \(n\). The reduction in variance is observed in the region where \(n\) is close to \(p\), and occurs in part because we tend to find unimodal predictive distributions in this region when we consider finitely many sample parameter candidates from a Gaussian distribution. It is possible that this shrinkage is an artifact of the experimental design, and Figure \ref{fig:optimized_pdfs} provides some evidence that when the prior puts weight on certain ``optimal'' parameters, this shrinkage does not occur. Figure \ref{fig:opt_pdf23} provides representative examples of predictive pdfs obtained following the setting of Section \ref{constructed} when \(n=p\). We see that there is no evidence of shrinkage as \(n\) increases, and most examples demonstrate multimodality.

\begin{figure*}
	\includegraphics[width=0.9\linewidth]{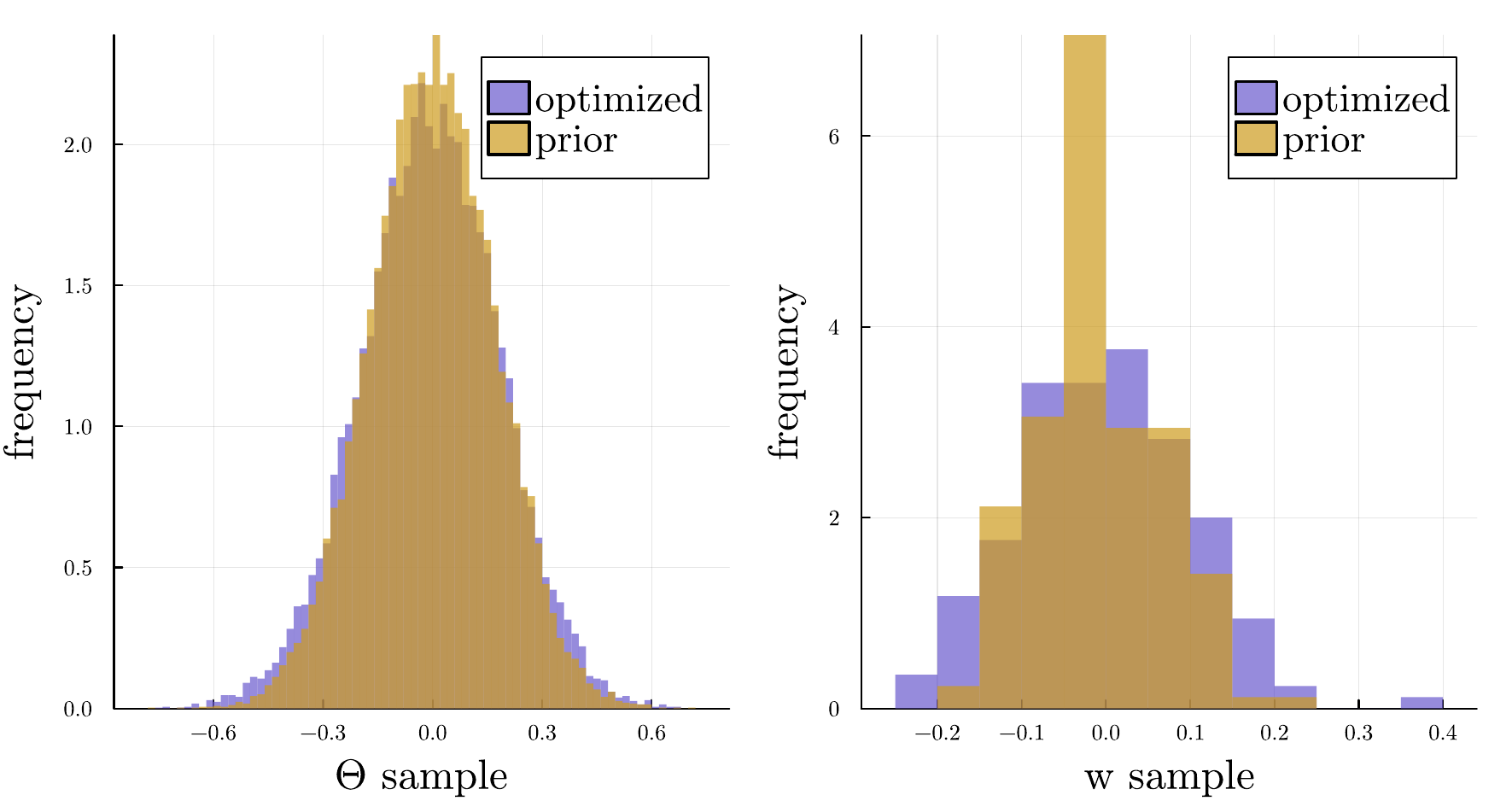} 
    \caption{\label{fig:prior_comparison} Comparison between the distribution of representative parameters constructed as described in Section \ref{constructed} (indigo) and parameters sampled from the prior used in Section \ref{gaussian} (gold). The left plot shows interior parameters, \(\Theta\), while the right plot shows final layer parameters, \(w\). }
\end{figure*}

Finally, it is worth examining the distribution of the components of the constructed parameter candidates, \(\Theta^{(j)}\), and their corresponding final layer weights, \(w\). If these constructed parameters are far outside typically used priors, the predictive modes they produce would not be informative about the behaviors of BNNs in practice. Figure \ref{fig:prior_comparison} provides a representative comparison between the distribution of constructed ``optimal'' parameters (indigo) to the distribution of the parameters drawn from the prior distributions considered in Section \ref{gaussian} (gold). We see that the distributions are close, but for both \(\Theta\) and \(w\), the variance of the distributions on the constructed parameters is slightly wider. Note that these comparisons do not capture the correlation between elements within \(\Theta\) of \(w\). Thus, it is reasonable that the constructed parameter candidates may be found when inference is performed with a Gaussian prior, but we may expect these candidates to behave differently than candidates with elements independently sampled from a Gaussian distribution.

\begin{figure*}
	\begin{tabular}{cc}
	\includegraphics[width=0.45\linewidth]{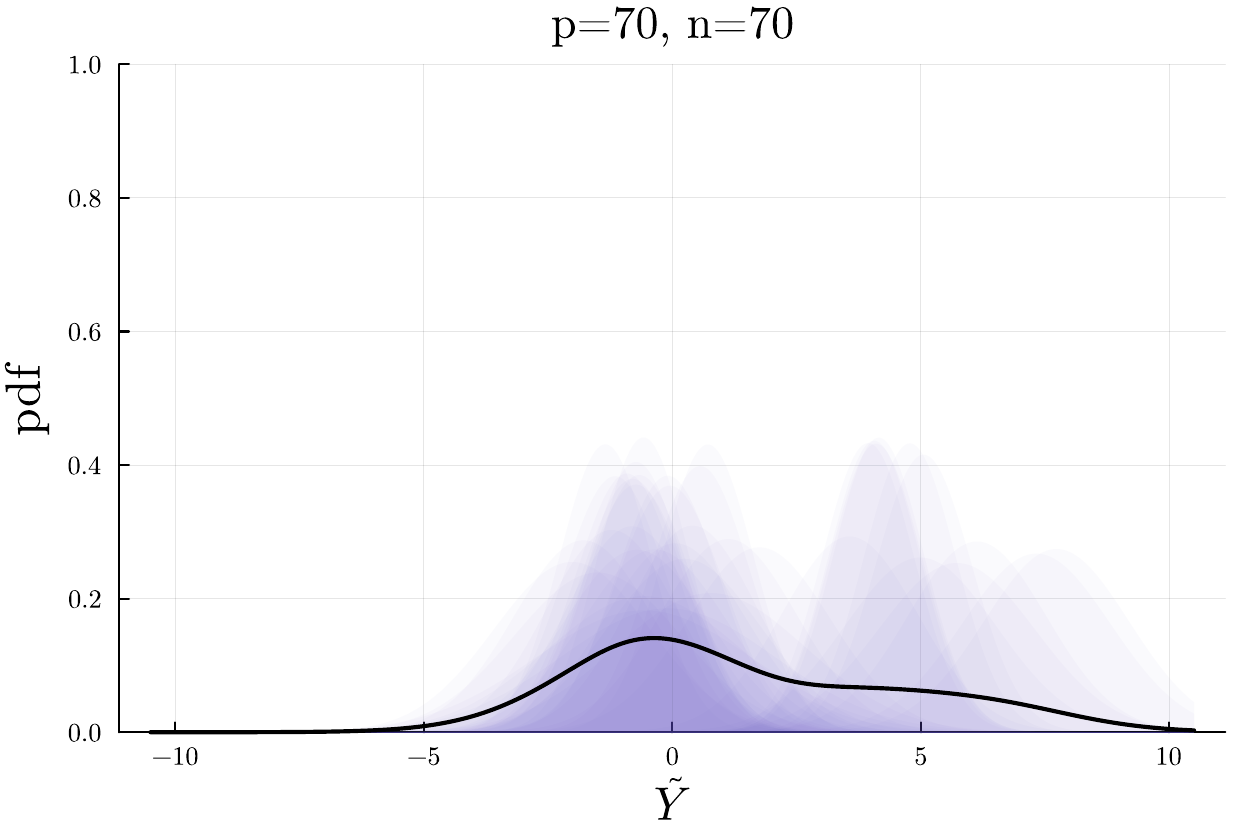} & 
	\includegraphics[width=0.45\linewidth]{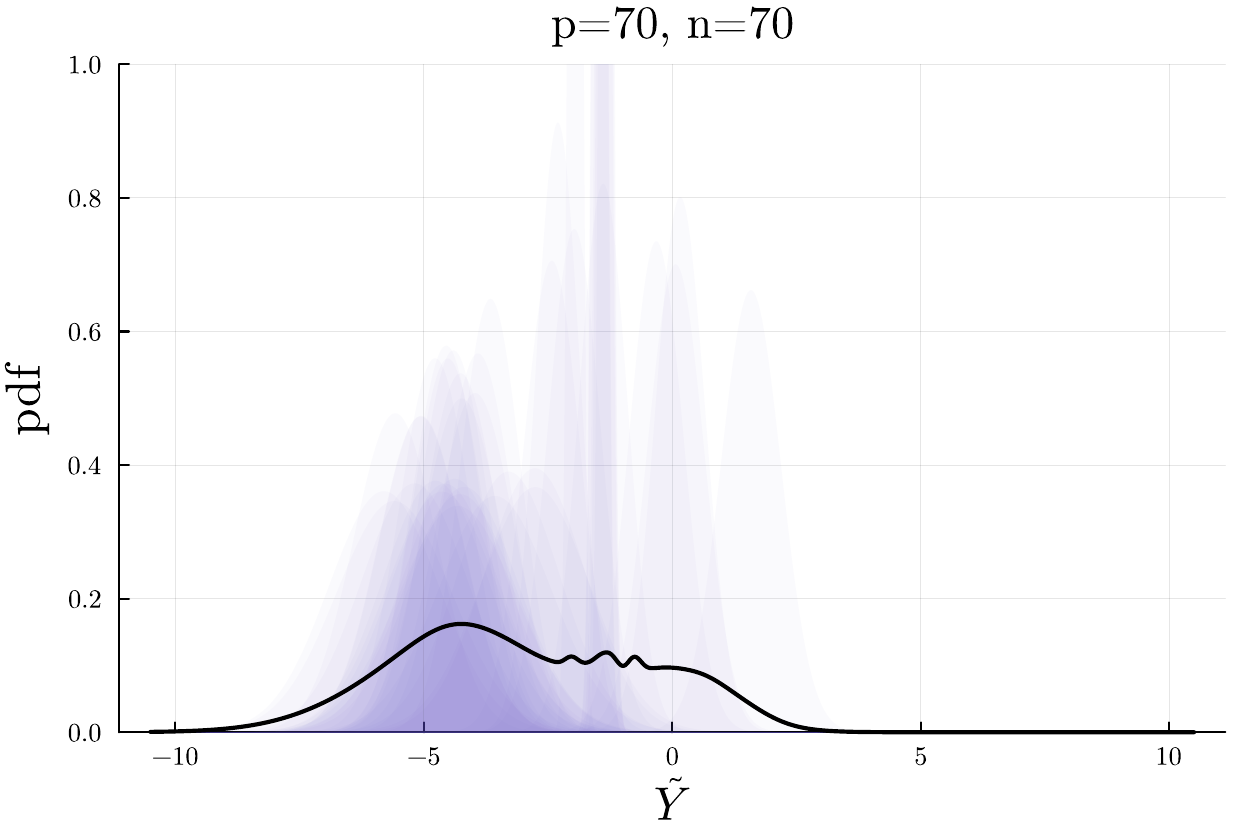} \\
	\includegraphics[width=0.45\linewidth]{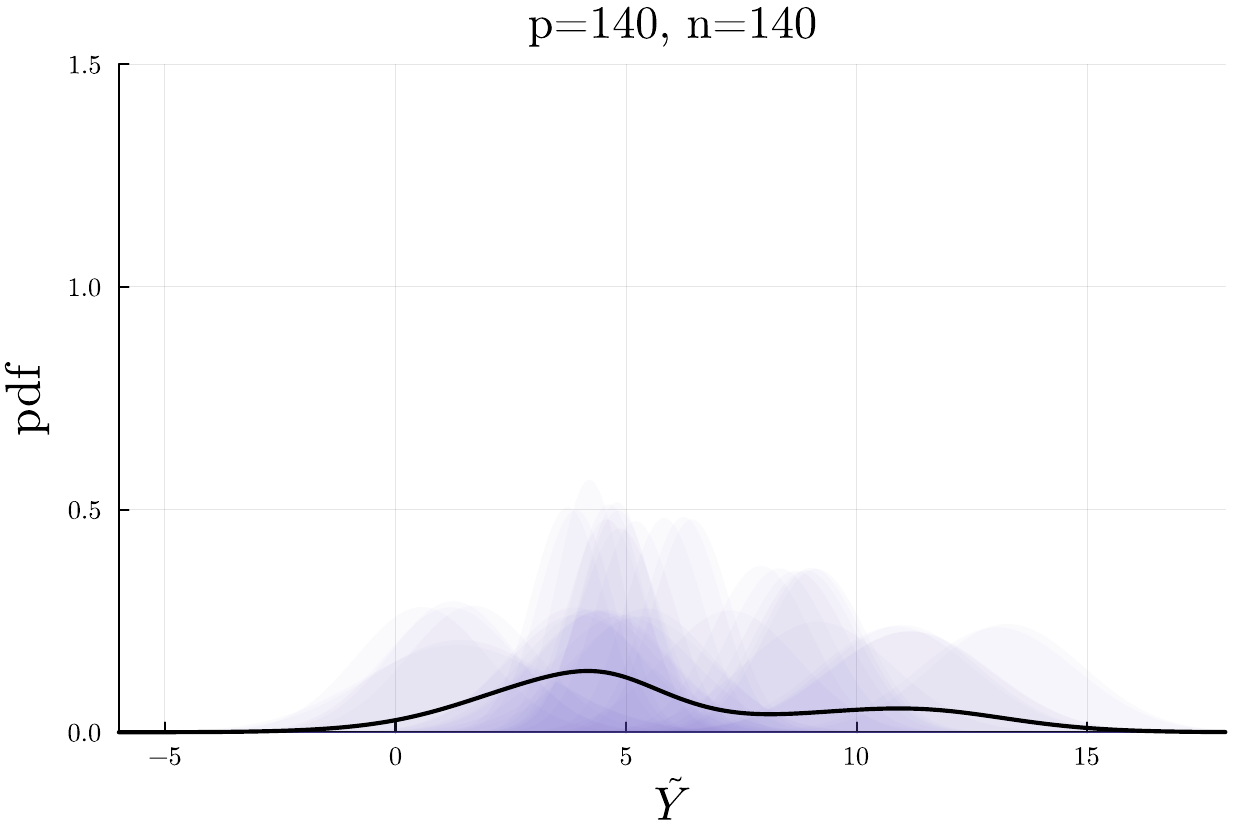} & 
	\includegraphics[width=0.45\linewidth]{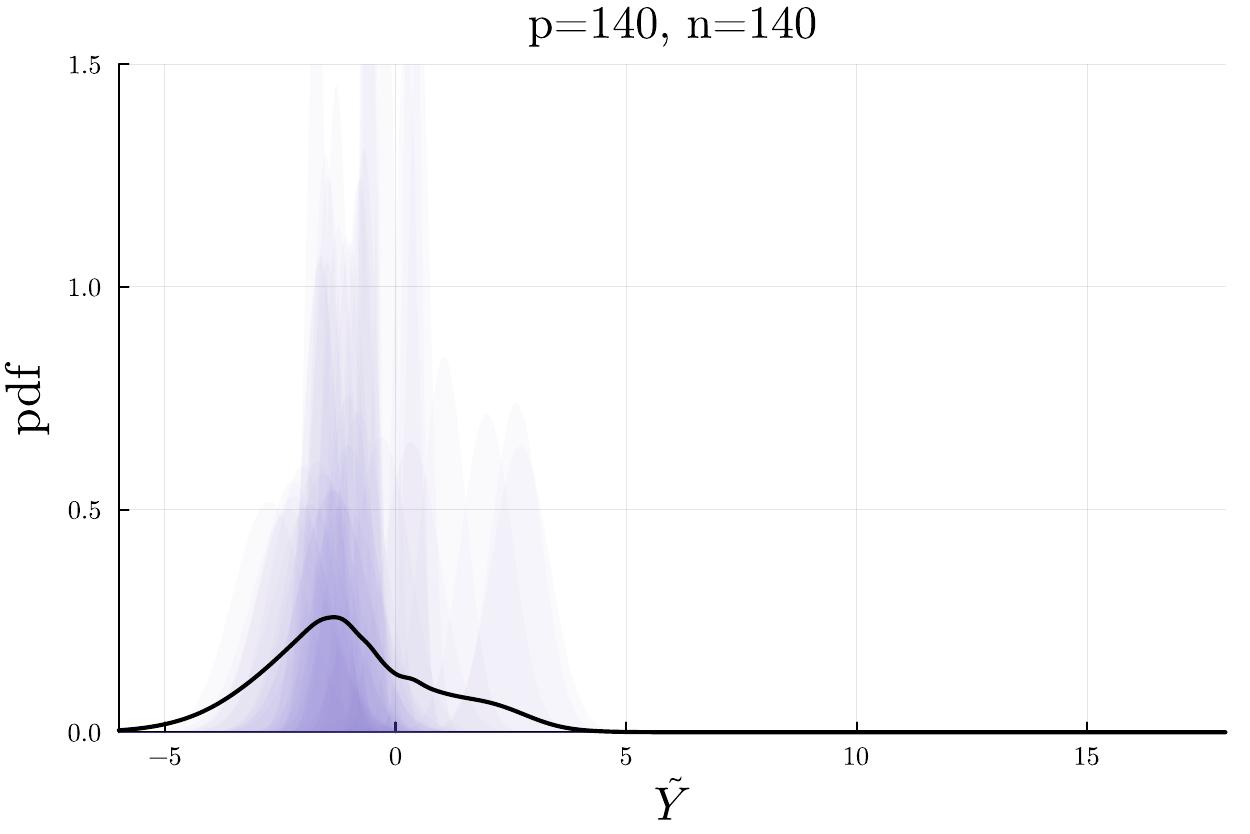} \\
	\includegraphics[width=0.45\linewidth]{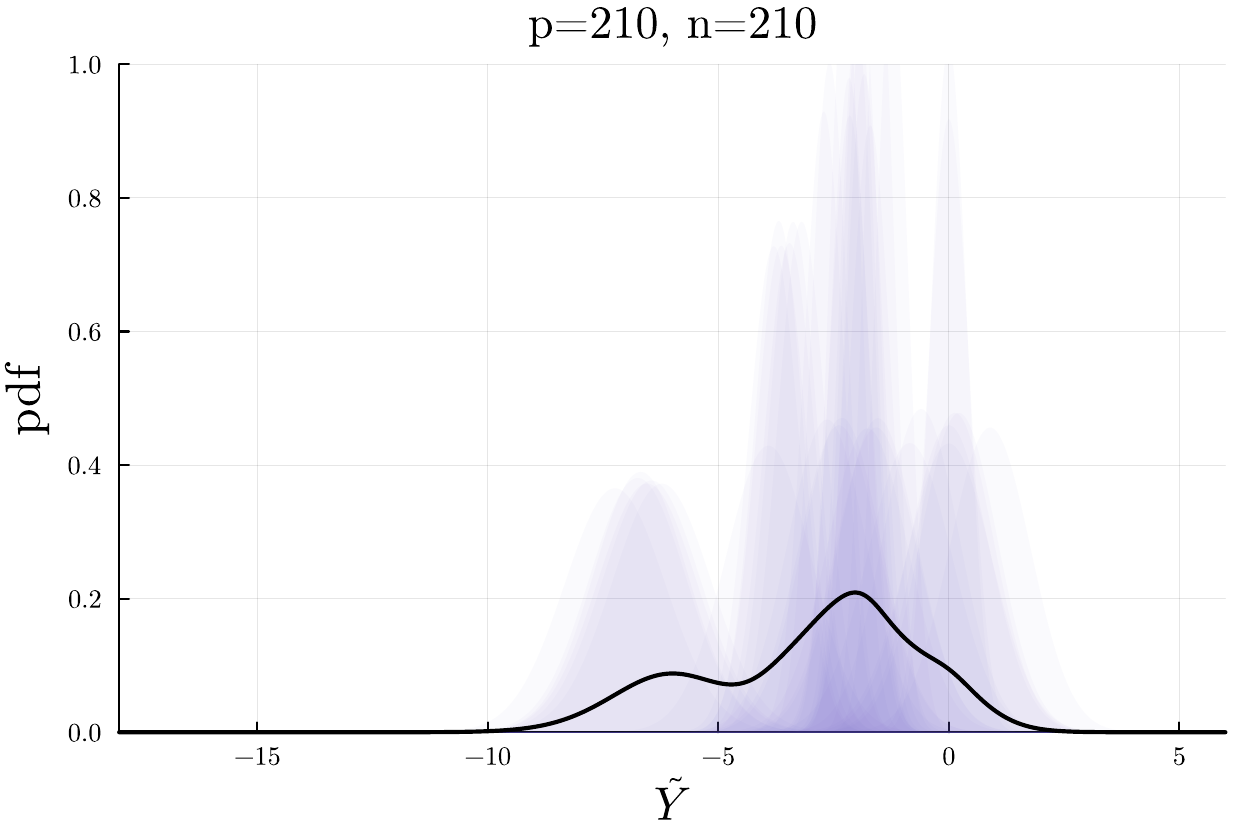} & 
	\includegraphics[width=0.45\linewidth]{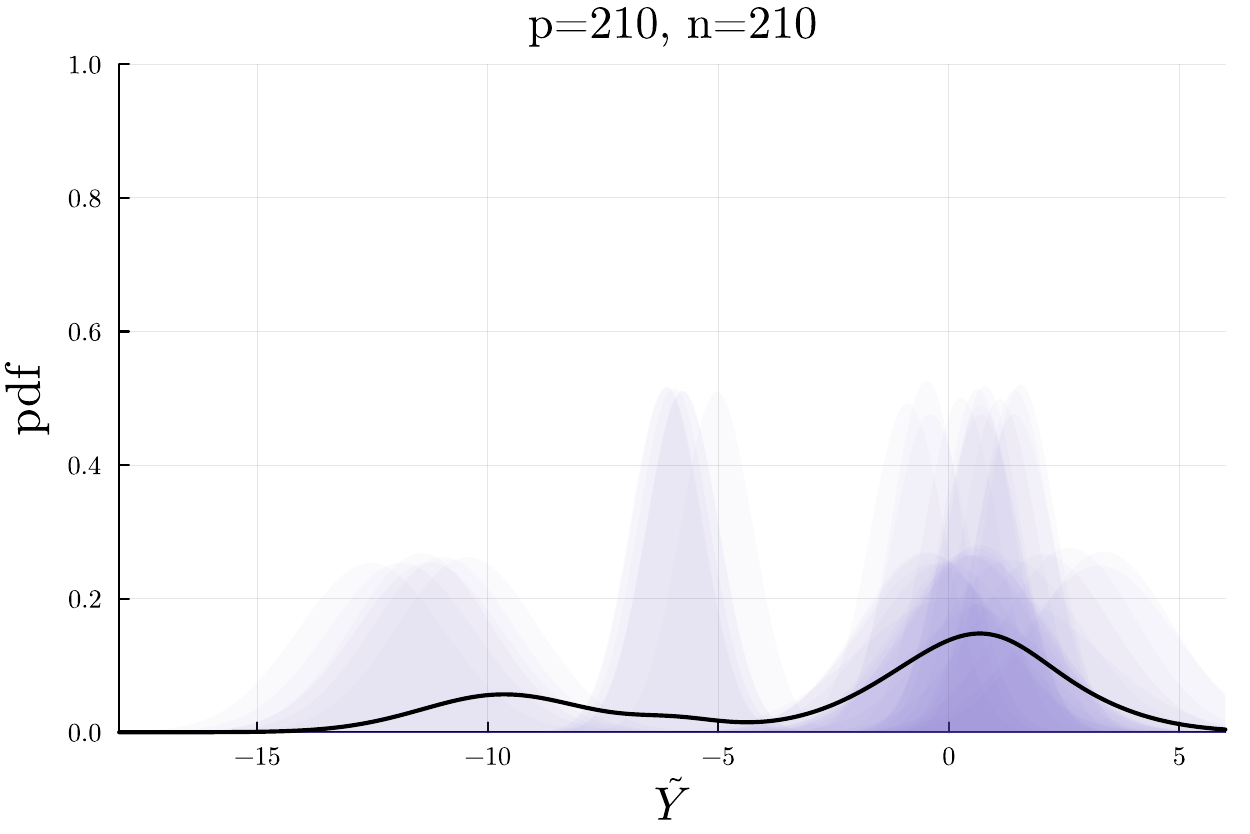} \\
	\end{tabular}
    \caption{\label{fig:opt_phase_pdf23} Predictive distributions for select \(n\) and \(p\) based on candidate parameters constructed to achieve \eqref{eq:conjecture}. The full distribution is plotted in black and components are shaded according to their weight in indigo. We consider \(10\) rotations, \(10\) preimage samples, and \(10\) column space samples to construct the distribution. Each row corresponds to a different input dimension; from top to bottom, we consider \(d\in\{100,200,300\}\).  Each column corresponds to a test location: \(\widetilde{x}_1^{(2)}\) (left) and \(\widetilde{x}_1^{(3)}\) (right).    }
\end{figure*}

\clearpage
\newpage


\end{document}